\documentclass[acmsmall]{acmart}
\usepackage{xurl}
\usepackage{makecell}
\usepackage{multirow}
\usepackage{array}
\usepackage{tabularx}
\usepackage{tikz}
\usepackage{forest}
\usetikzlibrary{positioning,arrows.meta,fit}

\newcommand{\yes}{\ding{51}}
\newcommand{\partc}{$\circ$}
\newcommand{\no}{\ding{55}}

\copyrightyear{2026}

\setcopyright{acmlicensed}
\acmJournal{CSUR}
\acmYear{2026}

\begin{document}

\title{Networked Humanoid Intelligence: Communication, Coordination,
  Collaboration, and Social Interaction}

\author{Loc X. Nguyen}
\orcid{0000-0001-5911-5847}
\email{xuanloc088@khu.ac.kr}
\affiliation{%
  \institution{School of Computing, Kyung Hee University}
  \city{Yongin-si}
  \country{Republic of Korea}
}

\author{Huy Q. Le}
\orcid{0009-0007-8342-7614}
\email{quanghuy69@khu.ac.kr}
\affiliation{%
  \institution{School of Computing, Kyung Hee University}
  \city{Yongin-si}
  \country{Republic of Korea}
}

\author{Ji Su Yoon}
\orcid{0009-0003-3928-1006}
\email{yjs9512@khu.ac.kr}
\affiliation{%
  \institution{School of Computing, Kyung Hee University}
  \city{Yongin-si}
  \country{Republic of Korea}
}
\author{Nguyen H. Tran}
\orcid{0000-0001-7323-9213}
\email{nguyen.tran@sydney.edu.au}
\affiliation{%
  \institution{The University of Sydney}
  \city{Sydney}
  \country{Australia}
}

\author{Eui-Nam Huh}
\email{johnhuh@khu.ac.kr}
\orcid{0000-0003-0184-6975}
\affiliation{%
\institution{School of Computing, Kyung Hee University}
\city{Yongin-si 17104}
\country{Republic of Korea}}

\author{Choong Seon Hong}
\orcid{0000-0003-3484-7333}
\email{cshong@khu.ac.kr}
\affiliation{%
  \institution{School of Computing, Kyung Hee University}
  \city{Yongin-si}
  \country{Republic of Korea}
}

\renewcommand{\shortauthors}{Nguyen et al.}

\begin{abstract}
Humanoid robots are moving from laboratory platforms toward connected fleets that share workspaces, infrastructure, tasks, and knowledge with other robots and humans. As deployments grow, the operation and improvement of each humanoid depend on information exchanged across the network. However, existing research has largely studied humanoids as individual systems. This survey examines networked humanoid intelligence through four functions of information exchange: communication, coordination, collaboration, and social intelligence with human interaction. We first establish the foundations of networked humanoid systems, define the question addressed by each function, and connect them through a common factory scenario. We review the literature along six dimensions: interaction architecture, exchanged information, physical coupling, operating timescale, system scale, and embodiment. We separate results obtained with physical and simulated humanoids from those based on other robotic platforms or disembodied agents, because the latter do not fully capture humanoid-specific constraints such as balance, whole-body motion, contact, and physical interaction. We examine hardware, datasets, and benchmarks, and distinguish risks handled locally from those networking introduces, such as stale information or unsafe knowledge propagation. Finally, we identify open problems in control-aware networking, compact body-centric messages, physically grounded coordination, network- and embodiment-aware knowledge sharing, fleet-level social consistency, and accountable human oversight.
\end{abstract}



\maketitle

\section{Introduction}
\label{sec:intro}

Humanoid robots are moving from laboratory prototypes toward large-scale deployment in manufacturing, logistics, service, and household environments. Major manufacturers are already preparing for mass production: Tesla is targeting a production capacity of up to 10 million Optimus robots~\cite{therobotreport2026optimus}, while a growing number of commercial humanoid platforms are entering industrial and service applications~\cite{hrt2026top12,sheng2025smartbot,cao2024humanoid}. This transition is supported not only by advances in humanoid hardware and embodied intelligence, but also by societal needs such as aging populations, labor shortages, and growing demand for care and service automation~\cite{yu2026ioh}. As deployment scales, humanoids will increasingly operate as part of larger populations: a factory may operate hundreds of robots at a single site, while a service provider may manage a much larger population distributed across homes and workplaces. Therefore, humanoids should be seen as members of connected populations that interact with other robots, shared infrastructure, and humans.

The rapid progress of humanoid robotics has largely focused on improving the capabilities of individual robots. Recent advances in reinforcement learning (RL) and whole-body control enable humanoids to walk, recover balance, track human motions, and perform complex whole-body behaviors~\cite{berkeleyhumanoid2024,fu2024humanplus,exbody2_2024,humanoidgym2024}. At a higher level, vision-language-action (VLA) models translate visual observations and language instructions into robot actions, enabling a single humanoid to perform diverse manipulation and loco-manipulation tasks~\cite{brohan2023rt2,kim2024openvla,black2024pi0}. World models further capture environment dynamics and predict action outcomes, supporting planning, policy evaluation, and behavior generation~\cite{hafner2025dreamerv3,worldmodelsurvey2025,ding2025worldmodelsurvey,humanoidworldmodels2025,dreamdojo2026}. In parallel, advances in visual perception improve state estimation and environment understanding~\cite{bin2025visualperception}, while motion-planning methods generate feasible actions under kinematic, stability, and real-time constraints~\cite{rutili2024motionplanning}. These developments provide the perception, prediction, reasoning, and control capabilities needed for increasingly autonomous humanoids. However, they primarily consider a single robot that observes its own environment and selects its own actions. They do not provide an explicit interface through which other robots can contribute information, coordinate decisions, or participate in joint control.

Consider two humanoids moving through a warehouse aisle while jointly handling a shelf module as a human worker passes nearby. The robots must first exchange task-relevant information in time to maintain a consistent view of the task, which is the role of \textbf{communication}~\cite{yu2026ioh}. Then, they have to decide who uses the aisle, who handles each part of the object, and when each action should occur, which requires \textbf{coordination}~\cite{choi2009cbba,multirobotsurvey2025}. When the shelf cannot be handled by either robot alone, they must synchronize their motions and forces and share useful experience across the fleet. These joint execution and learning processes form \textbf{collaboration}~\cite{hawley2019cooptransport,coophumanoidtransport,dualhumanoidtransport2024,ifl2023}. Finally, operating around people requires the robots to understand human intentions, communicate their actions, and behave safely and predictably, which requires \textbf{social intelligence and human interaction}~\cite{collabmind_intent2021,yu2026ioh}. These four capabilities define the main dimensions of networked humanoid intelligence.

\subsection{Why Humanoids Change the Networking Problem}
Although these capabilities appear in conventional multi-robot systems, humanoid embodiment imposes additional constraints on their operation. Bipedal balance and whole-body control rely on fast feedback, limiting the extent to which safety-critical control can depend on the network~\cite{yu2026ioh}. For example, a communication delay that may only interrupt the motion of a wheeled robot can disrupt synchronization between physically coupled humanoids, generate unwanted interaction forces, and threaten their balance~\cite{hawley2019cooptransport,coophumanoidtransport,dualhumanoidtransport2024}. At the same time, high degree-of-freedom (DoF) bodies and multimodal sensors generate large and heterogeneous data streams, making continuous exchange of raw observations impractical~\cite{bin2025visualperception,humanoideveryday2025}. Physical interaction creates another challenge. During cooperative manipulation, forces and motion errors can propagate through the manipulated object and disturb the balance of other robots~\cite{hawley2019cooptransport,coophumanoidtransport,dualhumanoidtransport2024}. Moreover, differences in morphology, DoF, sensing, and hardware limit the direct transfer of policies and experience across humanoids~\cite{oxe2023,hzero2025,crosshumanoidwbc2026}. This challenge becomes more important as humanoid knowledge is increasingly encoded in large learned policies and representations, whose exchange can require substantially more information than conventional task messages~\cite{groot2025,yuan2025survey,lwd2026,wang2023fleetmerge}. Therefore, networking for humanoids cannot be treated as an independent communication layer. Instead, it must be closely integrated with control, embodiment, and learning.

\subsection{Related Surveys and Positioning}

Existing surveys approach networked humanoid intelligence from several related directions. Surveys on humanoid robotics cover hardware design, locomotion and control, perception, planning, and foundation models, but primarily focus on the capabilities of individual robots~\cite{sheng2025smartbot,cao2024humanoid,rutili2024motionplanning,yuan2025survey}. Surveys on VLA and embodied models similarly examine perception-action architectures and language-conditioned control, with the robot typically treated as an individual decision-making agent~\cite{11495231,11164279,jiang2025survey}. By contrast, multi-robot and multi-agent surveys study communication, coordination, collaboration, multi-agent RL (MARL), and LLM-based teams~\cite{mrscollabtaxonomy2026,multiagentcoord2025,autonomousmrnav2024,gielis2022critical,llmmasurvey2024,wu2025generative,embodiedmasurvey2025}. However, these studies mainly consider wheeled robots, aerial robots, and manipulators, and therefore do not fully capture the physical and control constraints of humanoids. Networked-robotics surveys provide another complementary view by studying cloud, edge, IoT, and AIoT infrastructures for communication, computation offloading, resource management, and distributed intelligence~\cite{cloudrobotics2015,wu2025survey,connectedrobotics2026,zhang2025tsn}. The \emph{Internet of Humanoids}~\cite{yu2026ioh} considers connected humanoids, but primarily focuses on an architecture- and application-oriented view of the field. Therefore, a gap remains between studies of individual humanoid intelligence and studies of distributed robotic systems. In particular, existing surveys do not systematically organize the inter-agent capabilities that allow humanoids to exchange information, coordinate decisions, perform joint physical tasks, share learned experience, and interact with people under humanoid-specific constraints. This survey addresses this gap from a capability- and algorithm-oriented perspective. It organizes the literature around communication, coordination, collaboration, and social intelligence, and examines how these capabilities are implemented and evaluated on humanoid platforms.

\begin{table*}[t]
\centering
\caption{Comparison with representative surveys related to networked humanoid intelligence.
\yes: primary or systematic coverage;
\partc: partial or secondary coverage;
\no: not covered.}
\label{tab:survey_comparison}
\scriptsize
\setlength{\tabcolsep}{2.5pt}
\renewcommand{\arraystretch}{1.15}
\resizebox{\textwidth}{!}{%
\begin{tabular}{p{5.15cm} c *{9}{c}}
\toprule
\textbf{Survey} &
\textbf{Year} &
\makecell{\textbf{Humanoid}\\\textbf{specific}} &
\makecell{\textbf{Multi-robot}\\\textbf{/ networked}} &
\makecell{\textbf{Comm. stack}\\\textbf{\& real-time}} &
\makecell{\textbf{Coordination}\\\textbf{mechanisms}} &
\makecell{\textbf{Physical}\\\textbf{collaboration}} &
\makecell{\textbf{Collective /}\\\textbf{fleet learning}} &
\makecell{\textbf{Social}\\\textbf{\& HRI}} &
\makecell{\textbf{Network-scale}\\\textbf{safety / security}} &
\makecell{\textbf{Benchmarks}\\\textbf{\& datasets}}
\\
\midrule

Cloud robotics~\cite{cloudrobotics2015}
& 2015
& \no & \yes & \partc & \partc & \no & \yes & \partc & \partc & \partc
\\

Multi-robot communications~\cite{gielis2022critical}
& 2022
& \no & \yes & \yes & \partc & \no & \no & \no & \no & \no
\\

Humanoid motion planning~\cite{rutili2024motionplanning}
& 2024
& \yes & \no & \no & \no & \no & \no & \partc & \no & \no
\\

LLM-based multi-agent systems~\cite{llmmasurvey2024}
& 2024
& \no & \partc & \no & \yes & \no & \partc & \no & \partc & \partc
\\

General humanoid reviews~\cite{sheng2025smartbot,cao2024humanoid}
& 2025
& \yes & \partc & \partc & \no & \no & \no & \partc & \no & \no
\\

Humanoid visual perception~\cite{bin2025visualperception}
& 2025
& \yes & \no & \no & \no & \no & \no & \partc & \no & \no
\\

Autonomous and multi-robot navigation~\cite{autonomousmrnav2024}
& 2025
& \no & \yes & \partc & \yes & \no & \no & \no & \no & \no
\\

Multi-agent coordination~\cite{multiagentcoord2025}
& 2025
& \partc & \yes & \partc & \yes & \no & \partc & \partc & \no & \no
\\

Multi-robot collaboration systems~\cite{multirobotsurvey2025}
& 2025
& \no & \yes & \partc & \yes & \partc & \partc & \partc & \partc & \partc
\\

Embodied multi-agent collaboration~\cite{wu2025generative}
& 2025
& \no & \yes & \partc & \yes & \no & \partc & \partc & \no & \no
\\

VLA reviews~\cite{11164279,jiang2025survey,11495231}
& 2025-26
& \partc & \no & \no & \no & \no & \no & \partc & \no & \yes
\\

Cloud-edge-terminal collaborative intelligence~\cite{wu2025survey}
& 2025
& \no & \yes & \yes & \partc & \no & \partc & \no & \yes & \no
\\

Humanoid behavior foundation models~\cite{yuan2025survey}
& 2026
& \yes & \partc & \no & \partc & \no & \no & \partc & \no & \partc
\\

Multi-robot collaboration taxonomy~\cite{mrscollabtaxonomy2026}
& 2026
& \no & \yes & \no & \yes & \partc & \partc & \partc & \partc & \no
\\

Multi-agent embodied AI~\cite{embodiedmasurvey2025}
& 2026
& \no & \yes & \partc & \yes & \no & \partc & \partc & \partc & \yes
\\

AI-IoT-robotics integration~\cite{connectedrobotics2026}
& 2026
& \no & \yes & \yes & \partc & \no & \partc & \partc & \partc & \no
\\

Internet of Humanoids~\cite{yu2026ioh}
& 2026
& \yes & \yes & \yes & \partc & \partc & \yes & \yes & \yes & \no
\\

\midrule
\textbf{This survey}
& \textbf{2026}
& \yes & \yes & \yes & \yes & \yes & \yes & \yes & \yes & \yes
\\

\bottomrule
\end{tabular}
}
\vspace{-0.15in}
\end{table*}

\subsection{Contributions and Organization}
\label{sec:contributions}
The contributions of our survey are summarized as follows:
\begin{itemize}
\item We organize the literature into four functions, namely communication and networking, coordination, collaboration, and social intelligence and human interaction. These functions capture how connected humanoids exchange information, make joint decisions, share capabilities and experience, and reason about humans and other agents.
\item For each function, we explain why balance, whole-body contact, heterogeneous bodies, and human-facing motion make humanoids different from generic multi-robot systems, including a direct comparison of humanoid and generic robot communication. In addition, we explain how representative methods work and analyze them along common dimensions, distinguishing evidence from physical or simulated humanoids from results on other robots.
\item We assess available platforms and benchmarks, separate per-robot safety from network-induced risks such as stale information and fleet-wide propagation of unsafe knowledge, and derive open problems from the gaps identified in the review.
\end{itemize}

The remainder of this article is organized as follows. Section~\ref{sec:foundations} defines the four functions and the taxonomy. Sections~\ref{sec:communication}-\ref{sec:social} review communication and networking, coordination, collaboration, and social intelligence and human interaction, respectively. Section~\ref{sec:evaluation} examines evaluation, safety, and open problems, and Section~\ref{sec:conclusion} concludes.

\section{Foundations of Networked Humanoid Intelligence}
\label{sec:foundations}

This section introduces the core concepts used throughout the survey. We first examine the humanoid as a network node and explain how its control structure shapes its communication requirements.
We then distinguish communication, coordination, cooperation, collaboration, and social interaction according to their roles in a networked humanoid system. 
Section~\ref{sec:scenario} illustrates these functions through a common factory scenario, while Section~\ref{sec:taxonomy} presents the taxonomy and comparison dimensions.

\subsection{The Humanoid as a Network Node}
\label{sec:def}

A humanoid combines functions that operate at very different timescales, from high-level reasoning to fast whole-body control. These differences directly determine how each function can depend on the network. For example, GR00T~N1 runs a vision-language module (VLM) at 10~Hz and an embodiment-specific diffusion-transformer action module at 120~Hz~\cite{groot2025}. Helix similarly combines low-rate reasoning with a 200~Hz visuomotor policy and kilohertz joint control~\cite{figure2025helix}. Thus, planning, learning, and model updates can tolerate communication with an edge server, whereas state estimation and balance require fast local execution.

Humanoid balance places strict limits on network dependence. Unlike a wheeled robot, a humanoid must continuously maintain its balance even if communication is interrupted. During walking, its zero-moment point must remain within a support polygon that changes with each step~\cite{rutili2024motionplanning}. Therefore, a delayed control update can cause the humanoid to fall~\cite{zhu2020remotewbc,wagner2026races}. For communication flows that directly affect balance, tail latency and information freshness are more important than average throughput. In addition, humanoids produce large and heterogeneous data streams. For example, Humanoid Everyday records RGB, depth, LiDAR, tactile, inertial, and joint data at 30~Hz~\cite{humanoideveryday2025}. On the other hand, other robots need only task-relevant, body-centric information, such as active contacts  and upcoming footsteps. Knowledge sharing introduces a different challenge because learned policies are often tied to a particular embodiment, requiring alignment of joints and sensors before they can be transferred across robots~\cite{groot2025,hzero2025}. Communication also extends beyond robot-to-robot exchange: humans interpret a humanoid's gaze, posture, and motion as signals of intent~\cite{dragan2013legibility}. These differences mean that network dependence should be determined by the function being supported rather than by the humanoid as a whole~\cite{cloudrobotics2015,wu2025survey,yu2026ioh}. Fleet learning and balance control may operate on the same robot, but they require fundamentally different communication guarantees.

\subsection{Four Functions, Four Questions}
\label{sec:terms}

The terms communication, coordination, cooperation, and collaboration are used differently across the literature~\cite{mrscollabtaxonomy2026}. For example, task-allocation studies often use coordination to describe how tasks are assigned among robots~\cite{gerkey2004formal,korsah2013comprehensive}, whereas cognitive science characterizes joint action through shared representations and action prediction~\cite{sebanz2006joint}. To provide a consistent basis for this survey, we distinguish the four functions by the question each answers and the object on which it acts. Table~\ref{tab:functions} summarizes these distinctions.

\textbf{Communication and networking} considers \emph{what information is exchanged, with whom, when, and through which resources}. The exchanged information may include raw observations, learned features, state estimates, plans, gradients, or model parameters. Depending on the task, this information can be sent to another humanoid, a group, network infrastructure, or a human. Exchange can be periodic, event-triggered, or on demand, while the underlying resources include spectrum, routes, middleware, and computing infrastructure. Communication succeeds when the required information reaches its recipient while it is still useful~\cite{kaul2012real,9380899}. It enables information exchange but does not determine how the recipient should act.

\textbf{Coordination} concerns \emph{how decisions and actions are kept mutually consistent}. More specifically, it addresses  the \emph{compatibility among contributions} that each participant could otherwise perform independently. This compatibility can take several forms. \emph{Exclusion} prevents multiple humanoids from claiming the same task, tool, or space at the same time. \emph{Precedence} ensures that an action begins only after its prerequisite is completed~\cite{korsah2013comprehensive}. \emph{Synchronization} aligns actions that must occur together, while \emph{agreement} ensures that participants share the same decision, such as a selected leader or plan version. These constraints can be enforced through centralized assignment, auctions, consensus, priorities, conventions, or negotiation. Coordination succeeds when individual contributions remain compatible as the joint plan is executed.

Cooperation describes the intention of multiple agents to pursue a shared goal rather than compete with one another~\cite{mrscollabtaxonomy2026}. Unlike coordination, it defines the objective of the interaction. Therefore, cooperating humanoids may rely on communication, coordination, collaboration, or a combination of them. We assume cooperative participants throughout this survey and consider compromised or adversarial participants separately as a safety issue in Section~\ref{sec:eval:safety}.

\textbf{Collaboration} concerns \emph{how capabilities or experience are combined for joint execution or improvement}. It concerns the \emph{combination of contributions} to achieve an outcome that no humanoid can achieve alone, or can achieve as effectively. These contributions can be combined in four ways. \emph{Physical collaboration} combines forces through a shared object, where the resulting forces affect both object motion and the balance of the participating humanoids~\cite{wu2016symmetry}. \emph{Informational collaboration} fuses partial observations to construct a view that a single humanoid cannot obtain alone~\cite{li2022star}. \emph{Capability collaboration} combines complementary skills, tools, or physical abilities to perform a task~\cite{mandi2024roco}. Finally, \emph{experiential collaboration} shares data, policies, or learned experience so that knowledge acquired by one robot can improve others~\cite{ifl2023}. Across these forms, collaboration depends on how individual contributions are combined.

Collaboration requires coordination, but coordination does not necessarily lead to collaboration. For example, two humanoids carrying the same cockpit module must agree on their roles and lifting time (coordination) while combining their forces to move the module (collaboration). In contrast, two humanoids working at different stations may coordinate their tasks without combining their contributions. A simple removal test helps distinguish the two: if one participant is removed and the other can still complete its own part, the interaction is coordination; otherwise, it is collaboration.

\textbf{Social intelligence and human interaction} answers \emph{how human intentions, expectations, responsibilities, and control are incorporated into interaction}. It treats humans as participants with latent goals, social norms, and authority, rather than simply as obstacles or sources of information. This function cuts across the other three. Informing a worker is communication, yielding to the worker is coordination, and carrying a table together is collaboration~\cite{du2025cola}. These interactions become social when the humanoid must understand the human's intent, expectations, or authority to decide how to act.

From communication to collaboration, the dependence among agents becomes stronger, but the resulting network requirements differ. Physically coupled collaboration requires millisecond-level timing between robots, whereas experiential collaboration involves large data or model transfers that can tolerate much longer delays. Coordination instead depends on how quickly participants can reach and maintain agreement. Social interaction introduces another constraint: communication is limited not only by network resources but also by human attention.

\begin{table*}[t]
\centering
\caption{The four functions of networked humanoid intelligence compared on the same dimensions.}
\label{tab:functions}
\scriptsize
\setlength{\tabcolsep}{3pt}
\renewcommand{\arraystretch}{1.12}
\resizebox{\textwidth}{!}{%
\begin{tabular}{>{\raggedright\arraybackslash}p{1.9cm} >{\raggedright\arraybackslash}p{3.1cm} >{\raggedright\arraybackslash}p{3.0cm} >{\raggedright\arraybackslash}p{3.2cm} >{\raggedright\arraybackslash}p{3.2cm}}
\toprule
& \textbf{Communication and networking} & \textbf{Coordination} & \textbf{Collaboration} & \textbf{Social intelligence and human interaction} \\
\midrule
\textbf{Guiding question} & What information is exchanged, with whom, when, and through which resources? & How are decisions and actions kept mutually consistent? & How are capabilities or experience combined for joint execution or improvement? & How are human intentions, expectations, responsibilities, and control incorporated? \\
\textbf{Object} & Information flows and their resources & Compatibility among contributions & Combination of contributions & Humans' latent states, norms, and authority \\
\textbf{Exchanged content} & Observations, features, state, plans, parameters & Bids, roles, reservations, dependencies & Forces, gait phase, contact state, maps, skills & Intent estimates, legible motion, authority changes \\
\textbf{Timescale} & Control cycle to shift-level updates & Sub-second to minutes & Milliseconds (physical) to days (learning) & Sub-second (proximity) to weeks (teaming) \\
\textbf{Success} & Useful information arrives in time & Plan stays conflict-free during execution & Team achieves what no member can & People can anticipate, rely on, and override the fleet \\
\textbf{Factory example} & Heartbeats, obstacle events, nightly skill update & Station allocation, aisle reservation & Two humanoids lift the cockpit; a learned skill spreads & Handing a wrench; supervisor approves an exception \\
\textbf{Failure if absent} & Partners act on stale information & Conflicts, duplicated work, deadlock & Task infeasible; each robot relearns alone & People cannot predict or correct the robots \\
\textbf{Rep. studies} & \cite{li2022star,xue2026taskoriented,wagner2026races} & \cite{choi2009cbba,chen2025emos,alonso2019distributed} & \cite{hawley2019cooptransport,mcgill2011stretcher,lwd2026} & \cite{simtom_hrteams2025,dragan2013legibility,verhagen2024meaningful} \\
\bottomrule
\end{tabular}
}
\vspace{-0.15in}
\end{table*}

\subsection{One Deployment, Four Functions: A Factory Scenario}
\label{sec:scenario}

\begin{figure*}[t]
\centering
\includegraphics[width=0.9\textwidth]{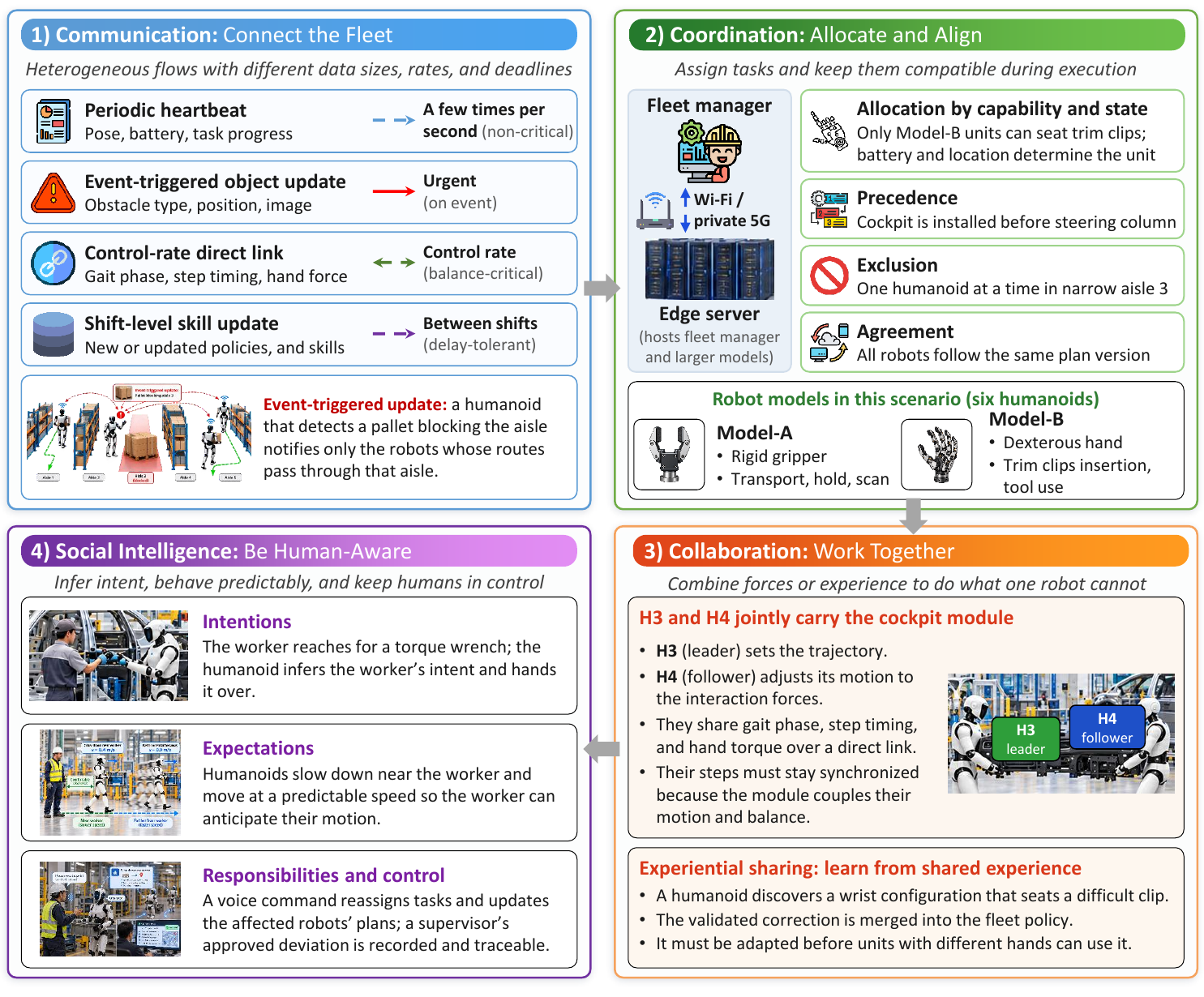}
\caption{Networked humanoids in an automotive assembly scenario~\cite{figure2025bmw}, where communication supports different types of information exchange~\cite{xue2026taskoriented,mcgill2011stretcher}, coordination assigns tasks and aligns robot actions~\cite{choi2009cbba,chen2025emos}, collaboration enables robots to combine physical capabilities and share experience~\cite{hawley2019cooptransport,lwd2026}, and social intelligence helps robots understand human intentions and maintain human oversight~\cite{hoffman2024intent,verhagen2024meaningful}.}
\label{fig:scenario}
\vspace{-0.2in}
\end{figure*}

Consider a segment of an automotive assembly line, illustrated in Figure~\ref{fig:scenario}. Humanoids of two models share the workspace: model-A units use grippers, whereas model-B units have dexterous hands. An edge server hosts the fleet manager and larger models, with connectivity provided through private cellular and Wi-Fi networks. A line worker operates at the door station, while a remote supervisor oversees several assembly segments. This scenario reflects the growing deployment of humanoids in industrial environments; for example, a Figure F.02 humanoid has accumulated more than 1,250 operating hours at BMW's Spartanburg plant~\cite{figure2025bmw}. We use this scenario throughout the survey to illustrate the four functions and their interactions.

\textbf{Communication.} Each humanoid periodically sends a compact status update, including its pose, battery level, and task progress, to the fleet manager. When a humanoid detects a pallet blocking an aisle, it sends an event-triggered update to the robots whose routes pass through that aisle. During cooperative transport, two humanoids exchange gait phase, step timing, and hand wrench at the control timescale over a direct link, where delayed or missing information can affect balance. Between shifts, larger skill updates are distributed across the fleet with much looser delay requirements. Therefore, the same humanoid generates communication flows that differ substantially in data size, frequency, and latency requirements.

\textbf{Coordination.} The fleet manager decomposes the assembly process into tasks and assigns the tasks according to humanoid capabilities and current states. For example, only model-B units can seat trim clips, while battery level and location affect which available humanoid receives a task. Coordination keeps these assignments compatible during execution. Precedence requires the cockpit to be installed before the steering column, exclusion allows only one humanoid at a time in narrow aisle~3, and agreement ensures that all participants follow the same plan version. 

\textbf{Collaboration.} Some tasks require multiple humanoids to combine their contributions. For example, the cockpit module is too long and heavy for one humanoid to handle alone, so two humanoids carry it together. The leader humanoid sets the motion trajectory, while the follower adjusts its motion according to the interaction forces. Their steps must remain synchronized because the shared module transfers the motion of each robot to the other and can affect their balance. Collaboration can also take the form of shared experience. When a humanoid discovers a wrist configuration that reliably seats a difficult clip, the validated correction can be incorporated into the fleet policy. However, the correction must be adapted before transferring to units with different hand configurations.

\textbf{Social intelligence.} The worker reaches for a torque wrench, and the humanoid infers his intent and hands the wrench to the worker autonomously. When humanoids move near the worker, they maintain a predictable speed so that the worker can anticipate their motion. Human instructions must also propagate through the fleet. If the worker reassigns a task by voice, the affected robots must update their plans accordingly. Similarly, when the remote supervisor approves a deviation from the current plan, the decision must be recorded and traceable.

Each function prevents a different type of failure. Without communication, two coordinating humanoids may lose the information needed to synchronize their motion. Without coordination, two humanoids may attempt to enter a limited-space aisle at the same time. Without collaboration, the fleet cannot complete tasks such as carrying the cockpit module that require combined capabilities. Without social intelligence, the fleet cannot reliably interpret human intent, respond to human instructions, or behave predictably around workers.

\subsection{Taxonomy and Review Dimensions}
\label{sec:taxonomy}

\begin{figure}[t]
\centering
\resizebox{\linewidth}{!}{%
\begin{tikzpicture}[
  font=\scriptsize,
  box/.style={draw, rounded corners=2pt, inner xsep=2.5pt, inner ysep=1.5pt},
  leaf/.style={box, text width=132pt, minimum height=19pt, align=left},
  sec/.style={box, text width=104pt, align=left},
]

\node[box, fill=gray!15, text width=40pt, align=center] (root) at (0pt,-78.4pt)
  {Networked\\humanoid\\intelligence};

\node[sec, fill=blue!6, anchor=east] (comm) at (-38.5pt,-40.25pt)
  {\textbf{Communication and}\\\textbf{networking} (Sec.~\ref{sec:communication})\\
   \emph{What to send, to whom, when,}\\\emph{and through which resources?}};
\node[sec, fill=orange!8, anchor=west] (coord) at (38.5pt,-40.25pt)
  {\textbf{Coordination} (Sec.~\ref{sec:coordination})\\
   \emph{How are decisions and}\\\emph{actions kept consistent?}};
\node[sec, fill=green!7, anchor=east] (collab) at (-38.5pt,-116.5pt)
  {\textbf{Collaboration} (Sec.~\ref{sec:collaboration})\\
   \emph{How are capabilities and}\\\emph{experience combined?}};
\node[sec, fill=violet!7, anchor=west] (social) at (38.5pt,-116.5pt)
  {\textbf{Social intelligence and}\\\textbf{human interaction} (Sec.~\ref{sec:social})\\
   \emph{How are intentions,}\\\emph{expectations, responsibilities,}\\\emph{and control incorporated?}};

\draw (comm.east)  -- (-30.5pt,-40.25pt) -- (-30.5pt,-116.5pt) -- (collab.east);
\draw (coord.west) -- ( 30.5pt,-40.25pt) -- ( 30.5pt,-116.5pt) -- (social.west);
\draw (root.west)  -- (-30.5pt,-78.4pt);
\draw (root.east)  -- ( 30.5pt,-78.4pt);

\node[leaf, anchor=east] (c1) at (-159.5pt, -9.5pt) {Carriers: connectivity, medium access, and middleware (\S\ref{sec:comm:carrier})};
\node[leaf, anchor=east] (c2) at (-159.5pt,-30.0pt) {Content, timing,\\and recipients (\S\ref{sec:comm:what})};
\node[leaf, anchor=east] (c3) at (-159.5pt,-50.5pt) {Placement: communication-\\computation co-design (\S\ref{sec:comm:offload})};
\node[leaf, anchor=east] (c4) at (-159.5pt,-71.0pt) {Communication inside\\the control loop (\S\ref{sec:comm:control})};
\foreach \n in {c1,c2,c3,c4} \draw (comm.west) -- ++(-6pt,0) |- (\n.east);

\node[leaf, anchor=west] (o1) at (159.5pt, -9.5pt) {Who does what:\\allocation and scheduling (\S\ref{sec:coord:allocation})};
\node[leaf, anchor=west] (o2) at (159.5pt,-30.0pt) {Agreeing on\\shared decisions (\S\ref{sec:coord:decision})};
\node[leaf, anchor=west] (o3) at (159.5pt,-50.5pt) {Spatial coordination (\S\ref{sec:coord:spatial})};
\node[leaf, anchor=west] (o4) at (159.5pt,-71.0pt) {Runtime conflict\\resolution and repair (\S\ref{sec:coord:replan})};
\foreach \n in {o1,o2,o3,o4} \draw (coord.east) -- ++(6pt,0) |- (\n.west);

\node[leaf, anchor=east] (l1) at (-159.5pt, -96.0pt) {Physically coupled\\joint execution (\S\ref{sec:collab:physical})};
\node[leaf, anchor=east] (l2) at (-159.5pt,-116.5pt) {Reasoning and\\capability composition (\S\ref{sec:collab:reasoning})};
\node[leaf, anchor=east] (l3) at (-159.5pt,-137.0pt) {Fleet learning and\\cross-embodiment transfer (\S\ref{sec:collab:learning})};
\foreach \n in {l1,l2,l3} \draw (collab.west) -- ++(-6pt,0) |- (\n.east);

\node[leaf, anchor=west] (s1) at (159.5pt, -96.0pt) {Intentions: social state\\and common ground (\S\ref{sec:social:intent})};
\node[leaf, anchor=west] (s2) at (159.5pt,-116.5pt) {Expectations: legibility,\\proximity, norms, and trust (\S\ref{sec:social:expect})};
\node[leaf, anchor=west] (s3) at (159.5pt,-137.0pt) {Responsibilities and control: teaming,\\teleoperation, and oversight (\S\ref{sec:social:resp})};
\foreach \n in {s1,s2,s3} \draw (social.east) -- ++(6pt,0) |- (\n.west);
\end{tikzpicture}}
\caption{Taxonomy of networked humanoid intelligence. The first level follows the guiding question of each function; the second level follows the subsections of Sections~\ref{sec:communication}-\ref{sec:social}.}
\label{fig:taxonomy}
\vspace{-0.2in}
\end{figure}
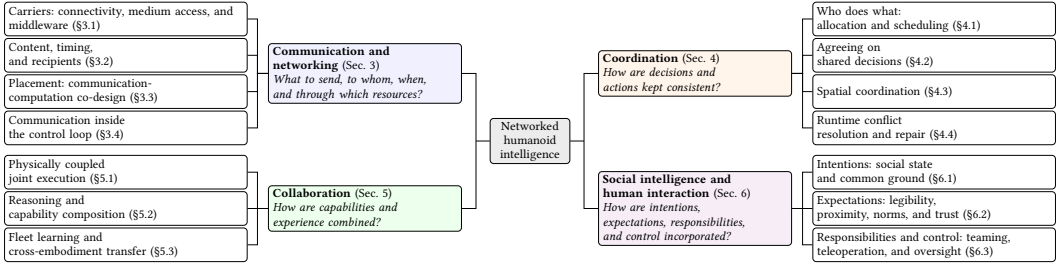

We organize the literature first by function, defined by the question each mechanism addresses, and then by the corresponding sub-questions as in Figure~\ref{fig:taxonomy}. Sections~\ref{sec:communication}-\ref{sec:social} follow these four functional branches. Existing taxonomies provide useful but narrower views: task-allocation taxonomies~\cite{gerkey2004formal,korsah2013comprehensive} focus on coordination, architecture-based classifications~\cite{multirobotsurvey2025} span multiple functions, and technology- or application-based classifications~\cite{wu2025generative,multiagentcoord2025} can separate mechanisms that address the same underlying problem. 

Within each branch, we analyze the literature through six dimensions: \emph{interaction architecture} (centralized, decentralized, or hybrid), \emph{exchanged information}, \emph{physical coupling} (none, shared space, or contact through an object or body), \emph{timescale} (control, reactive, planning, or learning), \emph{system scale}, and \emph{embodiment evidence}. We distinguish five levels of embodiment: physical humanoids (HR), simulated humanoids (HS), other physical robots (RR), other simulated robots (RS), and abstract agents (AB). This distinction is important because results from non-humanoid systems do not capture constraints such as balance, whole-body contact, and anthropomorphic interaction. The importance of each functional branch also depends on the deployment. For example, industrial humanoids~\cite{figure2025bmw} place greater emphasis on infrastructure coordination, whereas household humanoids~\cite{1x2026neo} require stronger social interaction and teleoperation-assisted autonomy.

\section{Communication and Networking}
\label{sec:communication}

\begin{figure*}[t]
\centering
\includegraphics[width=1\textwidth]{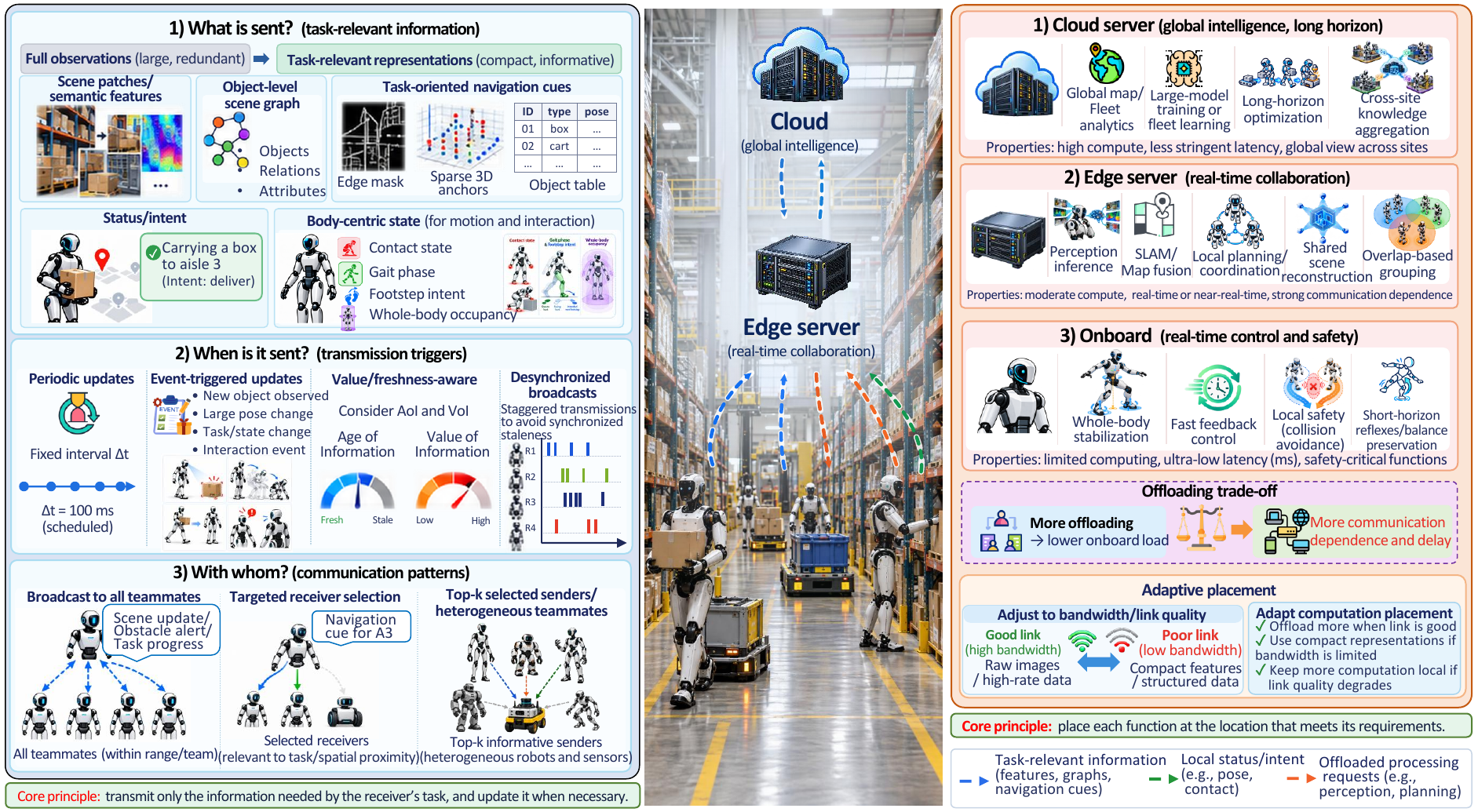}
\caption{Communication in networked humanoids: robots exchange compact, task-relevant information~\cite{li2022star,gu2025mrcographs} only when and to whom it is needed~\cite{xue2026taskoriented,das2019tarmac}, while computation is placed onboard, at the edge, or in the cloud according to its timing requirements~\cite{huang2022recslam,zhu2020remotewbc}.}
\label{fig:communication}
\vspace{-0.2in}
\end{figure*}

This section examines how communication supports networked humanoids operating as a group. The central problem is deciding \emph{what information should be exchanged, which humanoids should receive it, when it must arrive, and which communication and computing resources should carry it}~\cite{gielis2022critical,hong2019cloudswarm}. These decisions are closely coupled in humanoid systems. A message that is useful for fleet learning may tolerate seconds of delay, whereas information used for coordinated locomotion or manipulation can lose its value within a control cycle~\cite{dualhumanoidtransport2024}. Similarly, sharing richer sensor observations can improve a teammate's decisions but also increases bandwidth and processing demands~\cite{seraj2024hetero}. Therefore, we review how existing mechanisms select information, establish communication links, allocate network and computing resources, and adapt communication to the timing requirements of the physical task. Figure~\ref{fig:communication} illustrates two fundamental questions: what/when/who to send, and the placement of computation. The large volume of sensor data generated by a humanoid~\cite{humanoideveryday2025} creates a bottleneck both for network transmission and for onboard computation.
To address this, existing works replace continuous raw-data transmission with compact representations~\cite{gu2025mrcographs}, event-triggered~\cite{xue2026taskoriented}, or selective communication~\cite{seraj2024hetero}. To overcome limited onboard computing, several studies~\cite{rahman2019offloading,ijaz2026jcco} offload tasks to edge and cloud servers.

\textbf{Why humanoid communication is different.} Multi-robot networking often assumes robots that can stop safely, exchange low-dimensional states, and have simple physical footprints~\cite{gielis2022critical}. Humanoids challenge all three assumptions. First, communication deadlines are coupled with balance: a delayed packet used for whole-body control or cooperative tasks can cause instability or a fall~\cite{zhu2020remotewbc,wagner2026races}. Second, a humanoid generates heterogeneous traffic, ranging from control-rate proprioception to perception updates, language tokens, and model updates~\cite{groot2025,humanoideveryday2025}. Third, coordination requires body-centric information, such as contact state, gait phase, footstep intent, and whole-body occupancy, for which compact representations remain limited~\cite{liu2025ittakestwo,8624927}. Fourth, knowledge payloads can be large, while their usefulness depends on compatibility between the bodies of the sender and receiver~\cite{hzero2025}. Finally, humanoids also communicate intent to people through their motion, which should remain consistent with the information exchanged over the network~\cite{dragan2013legibility}. Table~\ref{tab:humanoid_vs_generic} summarizes these differences.

\subsection{Connectivity, Medium Access, and Middleware}
\label{sec:comm:carrier}

For humanoid fleets, communication resources are often constrained more by timing than by data volume. For example, streaming a 256-dimensional observation every 6.67~ms requires only about 1.3~Mbit/s per robot~\cite{wagner2026races}, yet this rate is already close to the usable throughput measured over an 802.11s mesh~\cite{chovet2025performance}. More importantly, the communication path must deliver these data with sufficiently low and predictable delay. We examine this requirement at three levels.

At the \textbf{substrate} level, infrastructure networks such as Wi-Fi, private cellular, and Open RAN provide centralized access and support traffic prioritization, whereas mesh networks remove the need for fixed infrastructure at the cost of less predictable delay~\cite{gielis2022critical}. Time-sensitive networking reduces this uncertainty on wired links by scheduling transmissions through synchronized switch gates~\cite{zhang2025tsn}, while digital twins can jointly model radio propagation and robot navigation~\cite{andrei2024digital}. However, neither approach has yet been demonstrated with mobile humanoid fleets. Connectivity can be incorporated into robot motion. Robots predict link quality at unvisited locations and choose paths that preserve connectivity~\cite{muralidharan2021communication}, while Mox et al.~\cite{mox2024opportunistic} deploy dedicated relay robots to improve the delivery probability of opportunistic routing. This strategy is less practical for humanoid fleets when every humanoid has a physical task and cannot simply act as relay nodes.

At \textbf{medium access}, 5G uplink transmission typically requires a scheduling request followed by a grant, adding delay and jitter before data can be sent.
O-RACES~\cite{wagner2026races} reduces this delay by predicting uplink demand. A lightweight DLinear forecaster at the Open RAN distributed unit uses the previous 100 traffic slots to predict demand 30 slots ahead, allowing grants to be issued before packets arrive. The predictor penalizes underestimation more heavily than overestimation because a missing grant delays transmission by another request cycle.

At the \textbf{middleware} level, ROS~2 communication latency varies with the DDS implementation, message size, and system configuration~\cite{maruyama2016exploring,park2020realtime,puck2021performance,gambo2025systematic}. ROS~2 also introduces additional overhead compared with raw DDS~\cite{kronauer2021latency}, while the best transport depends on the underlying network~\cite{zhang2024middleware}. This issue becomes more pronounced in dynamic meshes, where topology changes trigger costly DDS discovery traffic; Zenoh reduces this overhead through a lighter discovery mechanism~\cite{chovet2025performance}. At a slower planning timescale, agent protocols such as model context protocol and agent-to-agent support capability and information exchange between agents~\cite{ehtesham2025survey}. Despite these advances, evidence for humanoid fleets remains limited. Among the reviewed studies, only O-RACES directly relates a communication mechanism to physical outcomes such as humanoid falls, while none evaluates the mixed control, perception, language, and learning traffic generated by a humanoid.

\begin{table*}[t]
\centering
\caption{Differences between humanoid communication and generic multi-robot communication.}
\label{tab:humanoid_vs_generic}
\scriptsize
\setlength{\tabcolsep}{3pt}
\renewcommand{\arraystretch}{1.08}
\resizebox{\textwidth}{!}{%
\begin{tabular}{>{\raggedright\arraybackslash}p{1.6cm} >{\raggedright\arraybackslash}p{3.4cm} >{\raggedright\arraybackslash}p{4.8cm} >{\raggedright\arraybackslash}p{5cm}}
\toprule
\textbf{Dimension} & \textbf{Generic multi-robot} & \textbf{Networked humanoids} & \textbf{Implication for the network} \\
\midrule

Safe state on link loss
& Can brake, hover, or hold a statically stable state
& Must continuously maintain balance over a changing support region~\cite{rutili2024motionplanning}
& Balance-related traffic requires low tail latency and fresh information, with onboard control as fallback~\cite{zhu2020remotewbc} \\
\cmidrule(lr){1-4}

Most time-critical content
& Pose, velocity, and planned path
& Contact state, gait phase, interaction force, and intended footsteps~\cite{wu2016symmetry,liu2025ittakestwo}
& Body-state messages require compact representations and explicit validity periods \\
\cmidrule(lr){1-4}

Spatial footprint
& Often represented as a point, disc, or fixed polygon
& Whole-body occupancy changes with torso motion, swinging legs, and arm movement~\cite{8624927,10375191}
& Shared occupancy information must represent both body geometry and its evolution over time \\
\cmidrule(lr){1-4}

Physical coupling
& Interaction often occurs from a stable mobile or fixed base
& Physical interaction can directly affect the balance of both partners~\cite{hawley2019cooptransport}
& Coupled tasks require tightly synchronized communication with low delay variation \\
\cmidrule(lr){1-4}

Traffic per node
& Dominated by either perception or control traffic
& Control, perception, language, and model updates originate from the same humanoid~\cite{groot2025}
& Different traffic flows require different priorities, latency targets, and resource allocation \\
\cmidrule(lr){1-4}

Knowledge exchange
& Mainly maps, states, and compact policies
& Learned policies and skills can be large and specific to the robot's embodiment~\cite{hzero2025,oxe2023}
& Knowledge transfer must account for payload size, embodiment compatibility, and update verification \\
\cmidrule(lr){1-4}

Human-facing channel
& Intent is mainly conveyed through lights, screens, or sounds
& Gaze, gestures, posture, and motion can directly convey intent~\cite{dragan2013legibility}
& Networked decisions and human-visible behavior should convey consistent intent \\
\cmidrule(lr){1-4}

Outcome that matters
& Packet delivery, throughput, and latency
& Communication failures can cause falls, balance violations, or excessive interaction forces~\cite{wagner2026races}
& Communication should be evaluated by its physical consequences as well as network metrics \\
\bottomrule
\end{tabular}
}
\vspace{-0.18in}
\end{table*}

\subsection{Selective and Learned Information Exchange}
\label{sec:comm:what}

A reliable link does not determine what information should be sent. The mechanisms below follow a common principle: transmit only the information needed by the receiver's task, and update it when necessary. They differ mainly in what information they select and what triggers transmission.

\emph{What to communicate:} STAR~\cite{li2022star} retains a task-agnostic scene representation: the sender transmits only a subset of scene patches in each frame, while the receiver reconstructs the missing content with a masked autoencoder using information received from earlier frames. MR-COGraphs~\cite{gu2025mrcographs} represents the environment as an object-level 3D scene graph and compresses the semantic features of each node before transmission; the recovered features also support map merging. Xue et al.~\cite{xue2026taskoriented} further restrict communication to information needed for navigation, including an edge mask, sparse 3D anchors, and an object table. Although these methods target different tasks, they share a common shift from transmitting complete observations to transmitting only the scene information the receiver needs, which is the main focus of the semantic communication paradigm~\cite{11192484}.

\emph{When to communicate:} Xue \emph{et al.}~\cite{xue2026taskoriented} trigger communication only when a predefined event occurs, such as detecting a new target or observing a large change in pose. However, not every change has the same importance. Therefore, GSRC~\cite{wu2024gsrc} prioritizes queued commands using both age of information (AoI), which captures how stale a command is, and value of information (VoI), which measures its contribution to the desired trajectory. Communication timing must be coordinated across robots. The work~\cite{zhao2026desync} maintains each robot's broadcast rate but staggers their transmission times. This prevents synchronized updates from making the shared map stale at the same time for all robots, which can otherwise cause multiple robots to select the same exploration frontiers.

\emph{Who to communicate and whom to receive:} CommNet~\cite{sukhbaatar2016commnet} provides the baseline without selection: each agent receives the average hidden state of all teammates within a fixed communication range, so the same controller handles different team sizes but treats every sender equally. SchedNet~\cite{kim2019schednet} selects senders when the shared medium admits only $k$ transmitters: each agent learns an importance weight, and only the $k$ agents with the highest weights broadcast, typically one agent sending one or two 16-bit values per step. On the other hand, TarMAC~\cite{das2019tarmac} selects receivers: each message carries a signature and a value, and attention between the sender's signature and the receiver's query, learned from task reward alone, determines how strongly each receiver uses the message. Its messages (a 16-dimensional signature and a 32-dimensional value) are far smaller than the $224\times224$ image observations. CommNet, SchedNet, and TarMAC are evaluated using abstract agents. HetNet~\cite{seraj2024hetero} extends targeted communication to heterogeneous teams, where some robots sense the environment and others act only on their teammates' messages. Robots communicate only within range, and messages are encoded according to the types of both sender and receiver. Binary messages cut the bits per round by about 200 times compared with real-valued baselines. However, when noise grows with distance, performance drops sharply even at high signal-to-noise ratio, showing that communication still struggles with the changing links of mobile robots. These mechanisms improve communication efficiency, but their benefits depend on assumptions about what information the receiver needs and how it uses the message. In addition, most approaches are evaluated mainly with abstract agents over idealized communication channels, with limited validation on real embodied robots. None of the methods has been tested on humanoids or communicates humanoid-specific states such as contact state or motion intent.

\subsection{Communication-Computation Co-Design}
\label{sec:comm:offload}
Offloading reduces onboard computation by moving part of the workload to edge or cloud servers, but it also introduces communication delay. It is beneficial only when uplink transmission, remote processing, and downlink transmission together are faster than local execution~\cite{yang2026r2x}. This trade-off is particularly relevant to humanoids: a foundation model that runs at 11~Hz onboard can reach 32~Hz on a desktop GPU~\cite{wagner2026races}, while time-critical functions such as whole-body stabilization must remain onboard. Existing work increasingly treats computation placement as a function-specific decision. Huang et al.~\cite{huang2022recslam} found that onboard  simultaneous localization and mapping (SLAM) kept a robot's CPU above 83\% load. Therefore, they propose a robot-edge-cloud SLAM architecture that streams raw scans to edge servers, which build and fuse the maps before sending a single map to the cloud, and assigns robots with overlapping maps to the same server to balance load and shrink the uploaded map. This reduces latency by up to 39\%, but assumes stable robot-to-edge links.
Zhai et al.~\cite{zhai2021cloudroid} show that offloading each robot's computation independently can increase communication overhead in cooperative mapping. The maps computed in the cloud are returned to the robots for sharing, doubling wireless traffic. Cloudroid Swarm~\cite{zhai2021cloudroid} moves both computation and inter-robot information exchange to the cloud, reducing bandwidth usage on robot-to-cloud links by 57-80\% compared with conventional offloading.
In multi-hop networks, offloading decisions also depend on the communication path to the server. Hong et al.~\cite{hong2019cloudswarm} propose a game-theoretic method that jointly determines whether each robot should offload its computation and which neighboring robots should relay its data. In simulation, the method reduces the combined latency and energy cost, achieving a performance gain of about 1.5 times over local computing. However, it assumes that robots remain nearly stationary during offloading.
Rahman et al.~\cite{rahman2019offloading} further include robot motion, allowing robots to move toward access points with better bandwidth. In a warehouse task, this reduces energy by about 40\% compared with a single cloud-assisted robot, at the cost of more than twice the travel distance and an offline plan that cannot adapt to changing links. Therefore, computation placement, routing, and robot motion become coupled decisions.

Offloading processing to edge servers reduces onboard computation but increases responsiveness dependency on network conditions. To address this trade-off, Yang et al.~\cite{yang2026r2x} demonstrate FollowMe, a person-following robot that switches between JPEG-compressed images and compact vector-quantized features according to Wi-Fi signal strength to maintain reliable tracking.
Ijaz et al.~\cite{ijaz2026jcco} jointly optimize computation offloading, CPU frequency, transmit power, and bandwidth, reducing 41\% swarm energy consumption while meeting over 99\% of task deadlines. When the  bandwidth is limited, the method keeps more computation onboard to reduce transmission costs. However, it assumes that workloads can be divided between local and edge processing, which is suitable for perception tasks but less practical for tightly coupled balance-control policies.

\subsection{Communication Inside the Control Loop}
\label{sec:comm:control}
When networked information is used to control robot motion, communication delays can affect stability. For humanoids, one solution is to separate control functions according to their timing requirements. Zhu et al.~\cite{zhu2020remotewbc} apply this approach by splitting whole-body control between a 5G edge server and the robot. The server computes the expensive, slowly changing part of the controller, while the robot updates time-critical feedback locally using the latest received solution. When communication is delayed, the robot reuses the previous solution to maintain control. In simulations with a 37-DoF humanoid, this increases the maximum tolerable constant delay from 40 to 90~ms for balancing and from 3 to 25~ms for walking, compared with fully remote control.

Wen et al.~\cite{dualhumanoidtransport2024} take a different approach to address the delay issue. Because network fluctuations can desynchronize two NAO robots carrying a stretcher, the leader sends the follower a sequence of pre-planned commands, each tagged with an execution time, and both robots then execute the motion locally at the specified times. This avoids continuous communication during control, but requires the motion to be planned in advance and the robots' clocks to remain synchronized, and the paper does not measure network delay. Another approach is to reduce communication delays through network scheduling. Wagner and Wietfeld~\cite{wagner2026races} evaluate O-RACES using up to 16 simulated humanoids connected through a real Open RAN testbed. Compared with reactive scheduling, O-RACES reduces uplink latency by up to 50\%. During rough-terrain walking, reactive scheduling leads to humanoid falls, whereas proactive scheduling nearly eliminates them. These results show how network scheduling can directly affect humanoid stability.

\subsection{Summary and Lessons Learned}
\label{sec:comm:summary}

The existing studies address different aspects of communication. Connectivity and middleware support timely message delivery, selective exchange determines what information to share and with whom, computation placement determines which functions rely on the network, and control-loop designs mitigate the effects of delayed messages. These findings lead to four key lessons.

\textbf{Network requirements follow the consuming function, not the robot as a whole.}
A humanoid generates control, perception, planning, and learning traffic, and each flow has different timing requirements: balance control and physically coupled motion require fresh information with bounded delay; perception, planning, and model updates generally tolerate longer delays. For time-critical control, retransmitting an old packet can even be worse than dropping it since the information may no longer represent the robot's current physical state~\cite{yang2026r2x}. 

\textbf{The main bottleneck is increasingly information selection rather than raw capacity.}
Existing methods increasingly avoid transmitting complete observations and instead select compact latent features, semantic scene elements, event-triggered updates, or specific recipients~\cite{li2022star,xue2026taskoriented,wu2024gsrc,kim2019schednet}. AoI and VoI show that messages with the same size can have very different value depending on their freshness and relevance to the task. Humanoid fleets still lack compact representations for body-centric information such as contact state, gait phase, footstep intent, interaction force, and predicted whole-body occupancy, as well as how long this information remains valid.

\textbf{Communication, computation, and control placement should be designed together.}
Running a function remotely saves onboard computation but introduces network delay. Existing work jointly decides where to compute along with routing, resource allocation, and even robot motion~\cite{huang2022recslam,hong2019cloudswarm,ijaz2026jcco}, but their scenarios are limited to delay-tolerant tasks such as mapping and perception over stable links. Humanoid control places stricter requirements on computation placement. Slow functions can be offloaded, while time-critical feedback remains onboard to maintain control when remote results are delayed~\cite{zhu2020remotewbc}.

\textbf{Communication should be evaluated by its physical consequences.}
Most existing mechanisms are evaluated through throughput, latency, traffic volume, inference cost, or task success. Only a few studies relate network behavior to physical outcomes such as humanoid falls, and none examines how network impairment affects interaction forces during physically coupled tasks. Evaluation should connect delay, jitter, packet loss, and stale information to physical outcomes such as balance violations, task degradation, and interaction forces. Overall, communication provides timely and task-relevant information, but it does not ensure that different humanoids act consistently on that information, which leads to coordination.

\section{Coordination}
\label{sec:coordination}
\begin{figure*}[t]
\centering
\includegraphics[width=0.85\textwidth]{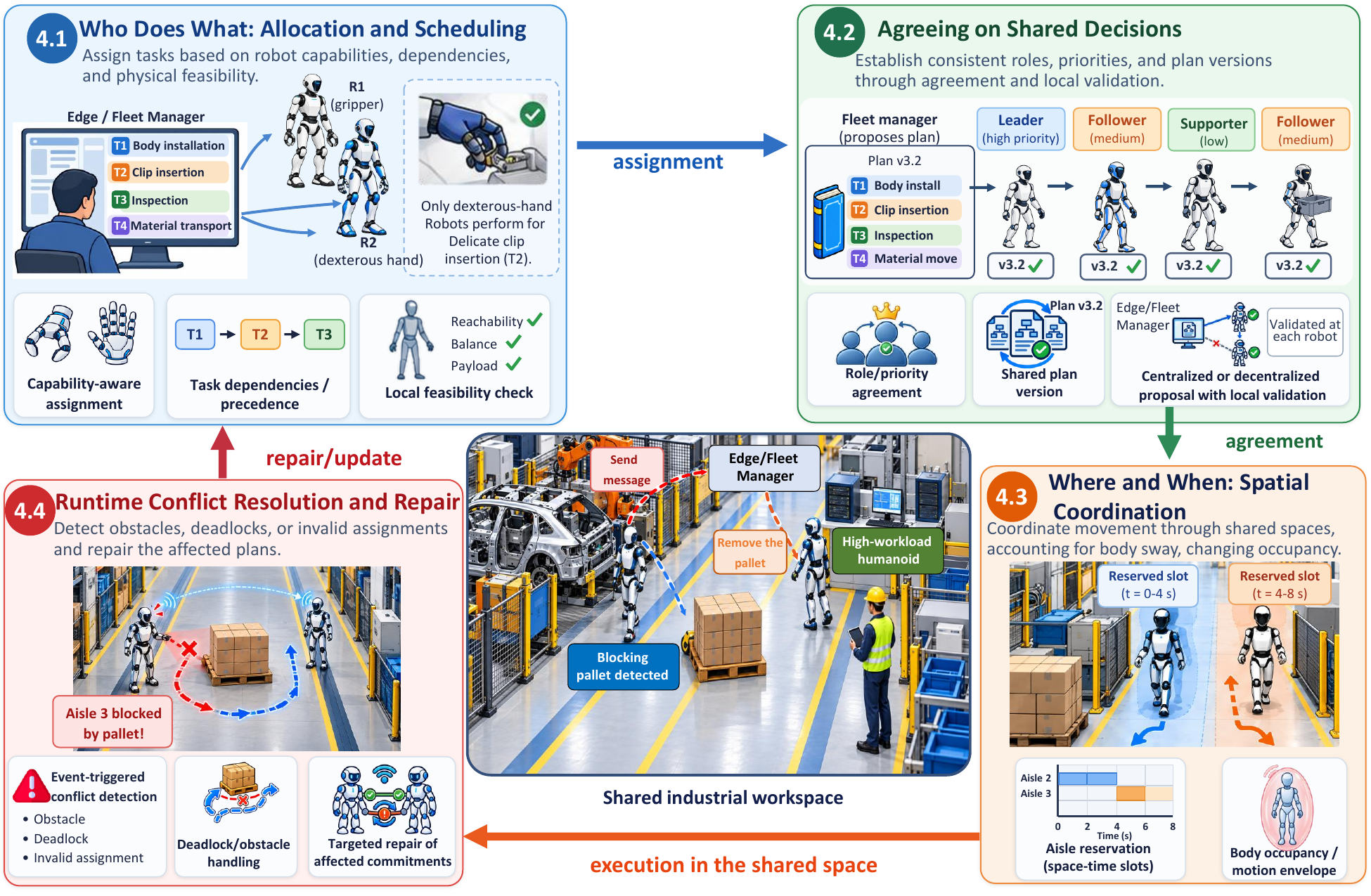}
\caption{Coordination in a shared workspace involves four processes: assigning tasks based on humanoid capabilities~\cite{chen2025emos,wang2025dartllm}, agreeing on roles and shared plans~\cite{chen2024scalablellm,liu2025coherent}, coordinating movement through shared spaces to avoid physical collision~\cite {alonso2019distributed,8624927}, and adjusting plans when unexpected events arise~\cite{huang2025compositional,garg2025foundation}.}
\label{fig:coordination}
\vspace{-0.15in}
\end{figure*}

Two robots with up-to-date information can still select the same task, block the same aisle, or execute dependent actions in the wrong order. Coordination addresses this problem by keeping the decisions and actions of multiple humanoids mutually consistent. This section reviews how networked humanoids enforce the exclusion, precedence, synchronization, and agreement constraints defined in Section~\ref{sec:terms} when coordinating actions that each robot could otherwise execute independently. Figure~\ref{fig:coordination} illustrates coordination as a loop of four steps in a shared workspace. First, tasks are assigned according to each robot's capabilities and task dependencies~\cite{chen2025emos,wang2025dartllm}. Second, the robots agree on roles and a shared plan version, often by letting a central planner propose the plan and each robot validate it locally~\cite{chen2024scalablellm,liu2025coherent}. Third, they share aisles and space by agreeing on collision-free regions and timing~\cite{alonso2019distributed,montano2016coordination}, while accounting for the body sway that changes how much space a humanoid occupies~\cite{8624927}. Finally, when a conflict occurs, such as a pallet blocking an aisle, the affected plans are repaired~\cite{huang2025compositional,garg2025foundation} and sent back for reallocation.

\textbf{Why humanoid coordination is different.} Humanoid coordination must account for physical constraints that change as the body moves. First, whether a humanoid can perform an assigned task depends on its current physical state: payload capacity varies with balance, contact forces are constrained, and carrying an object introduces fall risk that a simple utility score may not capture~\cite{chen2025emos,kannan2024smartllm}. Second, the space occupied by a humanoid changes with body sway, leg motion, and arm movement rather than remaining a fixed footprint~\cite{8624927,10375191}. Third, many coordination methods resolve conflicts by asking one robot to wait~\cite{cap2015prioritized,montano2016coordination}, but a humanoid cannot stop at every point in its motion and may need to step in place to remain stable. Finally, a humanoid may lose balance or deviate from its planned motion, requiring the team to adjust its actions during execution.

\subsection{Who Does What: Allocation and Scheduling}
\label{sec:coord:allocation}
Task allocation depends on how accurately the system represents what each robot can do. Classical methods represent this capability through a scalar utility for each robot-task pair~\cite{gerkey2004formal,korsah2013comprehensive,tao2026tamp}. For example, CBBA~\cite{choi2009cbba} allows each robot to build a sequence of tasks based on its expected benefits and exchange bids with its teammates to resolve conflicting assignments. This approach produces conflict-free assignments even when robots have different information about the environment. It also guarantees at least 50\% of the optimal reward when task values exhibit diminishing returns and robots have accurate information. However, a high utility score does not necessarily mean that a humanoid can safely perform the task unless its physical constraints are explicitly considered.

Capability-aware methods address this limitation by representing robot capabilities more explicitly.
EMOS~\cite{chen2025emos} uses robot descriptions in Unified Robot Description Format (URDF) and forward-kinematics tools to generate a ``robot resume'' describing mobility, perception, and manipulation capabilities, including limits such as arm reach. A central LLM assigns tasks, while each robot's LLM agent checks whether it can perform its assigned task and requests reassignment if necessary.
SMART-LLM~\cite{kannan2024smartllm} considers tasks that require multiple robots. It uses an LLM to divide tasks into subtasks, match them to predefined robot skills, and form teams when no single robot has the required capabilities. SMART-LLM with GPT-4 achieves 71\% success, compared with 14\% for rule-based allocation. However, its task assignments are planned before execution and are not updated as conditions change. 
Beyond assigning tasks to suitable robots, coordination must also consider the order in which tasks are performed. DART-LLM~\cite{wang2025dartllm} uses an LLM to divide instructions into subtasks and represent their dependencies in a directed task graph. This graph allows independent subtasks to run in parallel while ensuring that other subtasks follow the required order. In complex construction tasks, this improves the success rate of an 8B model from 45\% to 84\%.
Task assignments may also need to change when execution does not proceed as planned. COHERENT~\cite{liu2025coherent} addresses this problem by allowing individual robot agents to report their progress or failures to a central LLM, which uses the feedback to revise task assignments. In simulated multi-robot tasks, COHERENT achieves a 97.5\% success rate, compared with 75\% when execution history is removed.
Peng et al.~\cite{peng2025milp} combine LLMs with mathematical optimization for multi-robot task allocation. Locally deployed LLMs translate scheduling constraints, such as deadlines and task dependencies, into mixed-integer linear programming models, which a solver uses to assign and schedule robot tasks. 
Overall, these methods capture robot capabilities and task constraints but do not fully account for how a humanoid's physical state affects task feasibility.

Humanoid soccer provides an early example of considering robot motion in task allocation. Yi et al.~\cite{yi2016hierarchical} use each humanoid's walking planner to estimate how long it would take to reach the ball. Robots exchange these estimates, and the robot expected to arrive first becomes the attacker, with a mechanism to prevent frequent role switching. The approach was demonstrated in real RoboCup matches, but considers mainly locomotion time rather than broader physical constraints. Nadiri and Rad~\cite{nadiri2025swarm} take a broader approach by considering each humanoid's distance to the ball, kicking ability, and battery level when assigning roles. Their decentralized method uses ant-colony-inspired scores to update roles and adjust team formations as conditions change. It reduces role reassignment time from 5.1~s to 2.3~s compared with centralized coordination and maintains performance even when more than 50\% of communication packets are lost.

\subsection{Agreeing on Shared Decisions}
\label{sec:coord:decision}
Agreement mechanisms differ in what the team must agree on, where agreement is reached, and how much communication is required, as shown in Table~\ref{tab:coordination}. As discussed above, CBBA~\cite{choi2009cbba} reaches agreement by exchanging winning bids, winner identities, and update timestamps, without requiring robots to maintain identical information about their environment. However, the speed and reliability of agreement depend on the communication network. Olfati-Saber and Murray~\cite{1333204} show that stronger network connectivity can accelerate agreement but reduce tolerance to communication delays, revealing a trade-off between agreement speed and robustness. Shorinwa et al.~\cite{shorinwa2024distopt} review distributed optimization methods that allow robots to agree on shared solutions through local computation and communication. Their comparison shows that some methods tolerate unreliable or one-way communication but have difficulty enforcing constraints, while methods that handle constraints more directly often require reliable, two-way communication. To reduce the communication required for agreement, Epp et al.~\cite{otaconsensus2024} allow robots to transmit simultaneously, so their information is combined directly over the wireless channel. This enables robots to agree on a shared formation reference using only three orthogonal broadcast slots per update, regardless of team size.

Where agreement is reached also affects how well robots coordinate. Chen et al.~\cite{chen2024scalablellm} compare centralized, decentralized, and hybrid LLM-based planning. They find that decentralized LLM agents often spend many dialogue rounds repeating similar proposals, while a single central planner is more efficient but can overlook conflicts between robots' actions. Their hybrid approach allows a central planner to propose actions and individual robots to check their assignments and request revisions. This approach achieves the highest success rate across all four simulated tasks, although the additional feedback requires more LLM calls. Another approach is to resolve conflicts by assigning priorities to robots. Čáp et al.~\cite{cap2015prioritized} propose a decentralized method in which each robot plans its own path and adjusts it to avoid conflicts with higher-priority robots. This allows robots to coordinate without a central planner while considering their own motion constraints. Coordination can also occur without explicit negotiation. COLA~\cite{du2025cola} enables a humanoid to adapt its movements to a human partner while carrying an object together. It learns to interpret forces transmitted through the shared object from differences between joint positions and their targets, allowing the humanoid to follow human movements and reduce physical effort without dedicated force sensors. These studies show different ways to coordinate decisions, but few consider the physical constraints of humanoid robots. For networked humanoid fleets, reaching agreement is not enough. Each humanoid must also be able to carry out the agreed decisions safely as conditions change.

\begin{table*}[t]
\centering
\caption{Coordination mechanisms compared by what the team agrees on, where agreement is computed, the cost of one agreement round, and embodiment evidence.}
\label{tab:coordination}
\scriptsize
\setlength{\tabcolsep}{3pt}
\renewcommand{\arraystretch}{1.0}
\resizebox{\textwidth}{!}{%
\begin{tabular}{>{\raggedright\arraybackslash}p{2.4cm} >{\raggedright\arraybackslash}p{3.0cm} >{\raggedright\arraybackslash}p{3.4cm} >{\raggedright\arraybackslash}p{3.6cm} >{\raggedright\arraybackslash}p{1.0cm}}
\toprule
\textbf{Work} & \textbf{What is agreed} & \textbf{Where computed} & \textbf{Cost of one round} & \textbf{Evidence} \\
\midrule
CBBA~\cite{choi2009cbba} & Conflict-free task bundles & Decentralized consensus on bids & Bid vectors; rounds grow with network diameter & AB \\
EMOS~\cite{chen2025emos} & Capability-matched assignment & LLM agents discussing robot resumes & Tokens grow with team size & RS \\
DART-LLM~\cite{wang2025dartllm} & Task graph with dependencies & Central LLM; structured output & One decomposition & RS \\
COHERENT~\cite{liu2025coherent} & One subtask at a time & Central assigner, local action choice & One proposal-feedback cycle & RS \\
Hybrid dialogue~\cite{chen2024scalablellm} & Proposals subject to local veto & Central planner plus robot checks & One call plus local checks & RS \\
Soccer role selection~\cite{yi2016hierarchical} & Attacker, supporter, defender & Each robot's walking planner & One packet per second per robot & HR \\
Pheromone roles~\cite{nadiri2025swarm} & Roles with congestion penalty & Decentralized sampling & Local update per cycle & HS \\

Distributed optimization~\cite{shorinwa2024distopt} & Variables or gradients & One-hop neighbors & Repeated neighbor exchanges & RR \\
Over-the-air consensus~\cite{otaconsensus2024} & Consensus value & Computed in the wireless channel & One simultaneous transmission & AB \\
COLA~\cite{du2025cola} & Joint carry motion & Inside one proprioceptive policy & No messages; forces via the object & HS, HR \\
GAMECHAT~\cite{mahadevan2025gamechat} & Yield order at a passage & LLM dialogue; MPC with barriers & Dialogue only when conflict is near & RS \\
Deadlock escape~\cite{garg2025foundation} & Temporary leader and waypoints & Event-triggered LLM above a GNN & Invoked only on deadlock & RS \\
\bottomrule
\end{tabular}
}
\vspace{-0.15in}
\end{table*}

\subsection{Spatial Coordination}
\label{sec:coord:spatial}
Spatial coordination determines where robots move and how they safely share the same space.
One approach is to separate team-level planning from local motion control. Alonso-Mora et al.~\cite{alonso2019distributed} propose a distributed method in which robots exchange information with their neighbors to identify a shared collision-free region and adjust their formation to fit within it. Each robot then moves toward the assigned position while avoiding nearby obstacles. The method adjusted its formation to avoid a moving human, with a median planning time of 0.35~s and local control running at 20~Hz. However, the method assumes reliable communication without packet loss and simplified robot shapes. Its local planning can also lead to deadlocks. Maintaining a formation is more challenging when robots have different motion capabilities. Li et al.~\cite{hetformation_drl2026} address this problem with a car-like leader and omnidirectional followers. Their method combines physics-based control to account for the leader's motion, feedback control to correct tracking errors, and reinforcement learning to reduce the remaining errors. This improves formation accuracy compared with RL alone, although the method has only been evaluated in 2D simulation. While these approaches coordinate robots around predefined formation patterns, robots moving independently through shared spaces must instead avoid conflicts between their planned trajectories.
Proactive planning yields better paths only when these predictions are reliable, whereas reactive planning is safer but can be suboptimal~\cite{chung2024learning}. This trade-off motivates hybrid designs that combine long-horizon planning with fast local reactions~\cite{yang2025collision}.
Besides deciding where to move, robots must also coordinate when they move. Monta\~{n}o and Su\'{a}rez~\cite{montano2016coordination} propose an online method that allows robot arms to follow independently planned paths while adjusting the timing of their movements to avoid collisions. Each robot checks its teammates' upcoming positions and adjusts its progress without changing the original path. However, the method requires synchronized execution and may force robots to stop when collision checking cannot keep up. Applying this approach to walking humanoids would also require maintaining balance during such interruptions.

Humanoid embodiment introduces spatial constraints that position-based collision avoidance cannot capture. Existing humanoid soccer systems use position-based repulsion to avoid teammates~\cite{yi2016hierarchical,nadiri2025swarm}, but a humanoid's occupied space also changes with body sway and posture. Kumagai et al.~\cite{8624927} account for predicted center-of-mass sway when planning footsteps, then use these footsteps to guide collision-free whole-body motion planning within 82~ms per step. In simulation, this improves planning success through a narrow gate from 60\% to 86.7\%. Rossini et al.~\cite{10375191} address moving obstacles by adjusting upcoming footsteps in under 1~ms and using model-predictive control to maintain balance. Although both studies focus on single humanoids, they highlight the need to account for changing body occupancy during movement. For networked humanoids, compact \emph{motion envelopes} could represent these changes without exchanging complete joint trajectories.

\subsection{Runtime Conflict Resolution and Repair}
\label{sec:coord:replan}
A coordination plan remains valid only while its assumptions hold, so robots must detect and repair conflicts during execution. Because language models are too slow to run inside fast control loops, most methods call them only when a failure or conflict occurs, while a fast local layer keeps the robots safe in the meantime~\cite{li2026llmmrs}. 
At the task level, these methods differ mainly in when they catch a failure. RoCo~\cite{mandi2024roco} checks task feasibility and collisions before execution and uses failed checks to revise plans. 
COHERENT~\cite{liu2025coherent} further distinguishes between failures caused by an incorrect subtask, an unsuitable robot assignment, or an action that cannot be executed. This allows the central planner to identify what went wrong and adjust its next proposal accordingly.
Some failures also require closer analysis after execution. REFLECT~\cite{liu2023reflect} summarizes a robot's visual, audio, and state observations to identify what went wrong. It first checks whether each subgoal was completed, then examines relevant observations to explain execution failures or reviews the original plan to identify planning errors. These explanations guide recovery plans that succeed in about 80\% of simulated cases. However, the method focuses on single-robot tasks, assumes a static environment, and is less effective at diagnosing low-level control failures. 

Runtime conflicts can also arise during robot movement, even when collisions are avoided. Garg et al.~\cite{garg2025foundation} consider robot teams navigating around obstacles while maintaining communication connectivity. These constraints can cause the entire team to become stuck before reaching its goals. To resolve such deadlocks, their method uses a foundation model to suggest temporary waypoints for a leader, allowing the robots to move together out of the blocked region. A distributed graph-based controller guides their movements while maintaining collision avoidance and connectivity. Other motion conflicts can be resolved before robots become stuck. GAMECHAT~\cite{mahadevan2025gamechat} addresses conflicts when robots approach the same narrow passage. The robots use LLMs to exchange task information and agree on which robot moves first. The lower-priority robot has to adjust its speed to give way, while a safety-constrained controller prevents collisions. If the robots cannot reach an agreement, a game-theoretic rule determines the passing order.

\subsection{Summary and Lessons Learned}
\label{sec:coord:summary}
Table~\ref{tab:coordination} shows that coordination mechanisms differ in what the team must agree on, where agreement is reached, and how much communication and computation each decision requires. Across these mechanisms, four lessons emerge:

\textbf{Slow global reasoning is paired with fast local safety.} Across allocation, agreement, spatial coordination, and recovery, an effective design uses a slower global process to propose decisions and a fast local layer to check and execute them~\cite{chen2024scalablellm,liu2025coherent,alonso2019distributed,garg2025foundation}. The global process considers the whole team, while the local layer enforces constraints that are difficult to capture centrally. However, current local checks mainly cover feasible actions, kinematic reachability, and collision avoidance~\cite{liu2025coherent,mandi2024roco}, but rarely consider humanoid-specific constraints.

\textbf{Shared coordination state should better reflect humanoid physics.} Current teams mainly exchange bids, roles, positions, and target regions, while multi-humanoid systems often model each robot as a simple disc for collision avoidance~\cite{yi2016hierarchical,nadiri2025swarm}. However, single-humanoid planning shows that the space a humanoid needs changes with body sway, posture, and footsteps~\cite{8624927,10375191}. Therefore, coordination should include a compact \emph{motion envelope} that captures expected body occupancy, planned footsteps, allowable deviation, and how long this information remains valid.

\textbf{Agreement cost should be considered together with how long a decision remains useful.} Existing studies measure coordination cost in rounds, transmissions, tokens, but rarely connect this cost to physical execution. Agreement becomes less tolerant to delay as connectivity increases~\cite{1333204}, while language-based coordination may require several model calls~\cite{chen2024scalablellm} or seconds of dialogue~\cite{mahadevan2025gamechat}. Some methods resolve conflicts by making robots wait~\cite{montano2016coordination}, which is not always possible.

\textbf{Coordination state needs explicit validity and targeted repair.} 
Task assignments and space reservations can become outdated as the robot or network state changes. Existing methods address this by reducing confidence in shared estimates when they are not updated~\cite{coltin2010multi} and by revising only the tasks affected by a trigger~\cite{huang2025compositional}. This is important for humanoids because a stumble may invalidate only part of a plan while the robot remains active. Shared state should record the source, scope, and validity period of each piece of information, so that only the affected commitments need to be updated.

\section{Collaboration}
\label{sec:collaboration}
\begin{figure*}[t]
\centering
\includegraphics[width=0.70\textwidth]{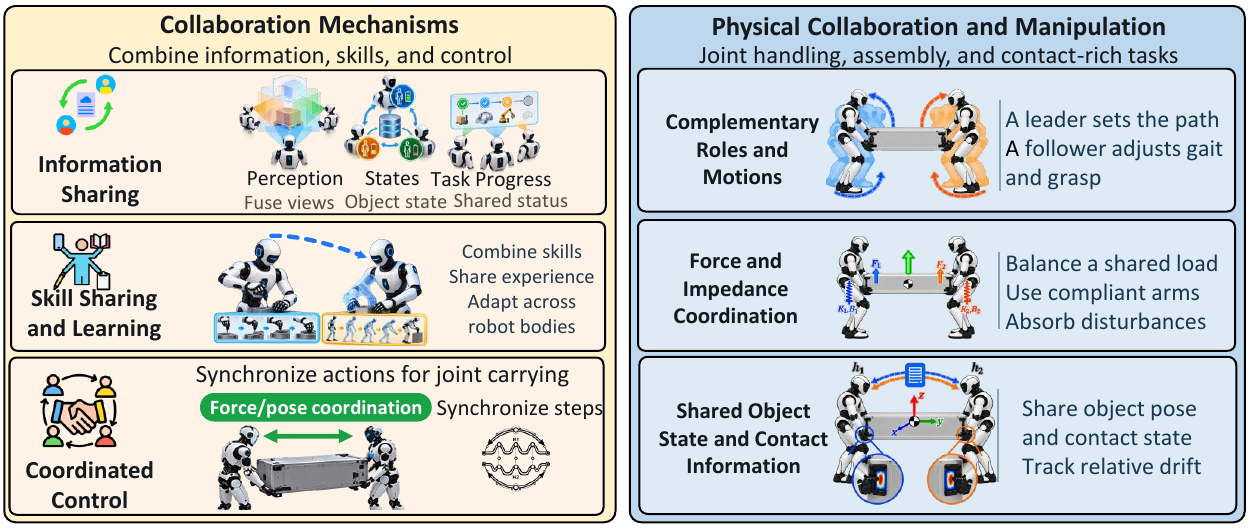}
\caption{Collaboration in networked humanoids: they share information~\cite{coela2023}, skills, and experience~\cite{lwd2026,hzero2025} and synchronize their actions~\cite{mcgill2011stretcher}. When carrying the same object~\cite{gasbarrone2024decentralized}, they use leader-follower roles, compliant arms~\cite{hawley2019cooptransport}, and shared contact state~\cite{liu2025ittakestwo}.}
\label{fig:Collaboration}
\vspace{-0.22in}
\end{figure*}

Coordination keeps individual contributions compatible; collaboration combines them to achieve outcomes that individual humanoids cannot achieve alone. For example, two humanoids may agree on when and where to lift a cockpit module, but the task still requires them to combine their physical capabilities during execution. This section reviews how networked humanoids combine capabilities and experience for joint execution and improvement. We consider four forms of collaboration: physical, informational, capability, and experiential composition. Figure~\ref{fig:Collaboration} illustrates two levels of collaboration: the general mechanisms that combine information, skills, and control, and the physical collaboration needed when humanoids handle the same object. A single humanoid has limited views, skills, and payload~\cite{coophumanoidtransport}, so many tasks cannot be completed alone. To address this, various works let humanoids share information to build a common view of the task~\cite{coela2023}, share skills and experience across robots with different bodies~\cite{lwd2026,hzero2025}, or synchronize their actions for joint carrying~\cite{mcgill2011stretcher}. In physical collaboration, the shared object couples the robots' balance, so existing studies assign leader and follower roles~\cite{gasbarrone2024decentralized}, use compliant arms to absorb disturbances~\cite{hawley2019cooptransport}, and exchange partner and contact state to track relative drift~\cite{liu2025ittakestwo}.

\textbf{Why humanoid collaboration is different.} Humanoid embodiment makes collaboration more tightly coupled than in wheeled robots or fixed-base manipulators. First, a shared object transfers one robot's motion and forces to its partner, directly affecting both robots' balance. Independent stabilization can produce conflicting corrections and increase interaction forces~\cite{wu2016symmetry,hawley2019cooptransport}. Second, interaction forces change every control cycle, whereas a humanoid may need to wait until the next step to correct its relative position~\cite{gasbarrone2024decentralized}. Third, successful collaboration requires the robot to treat its partner's actions as useful information rather than as disturbances to reject~\cite{liu2025ittakestwo}. Finally, sharing experience is particularly valuable since collecting humanoid data is costly and can involve physical risk, but differences in joints, links, and end effectors limit direct transfer across platforms~\cite{hzero2025,oxe2023}. Shared experience can propagate both useful skills and errors across the fleet.

\subsection{Physically Coupled Joint Execution}
\label{sec:collab:physical}
Large or heavy objects may require multiple humanoids to carry them together~\cite{coophumanoidtransport,hawley2019cooptransport}. However, carrying the same object physically couples their movements, so one robot can disturb the balance of another. This becomes particularly challenging in leader-follower control, where the follower adjusts its walking motion only after sensing forces caused by the leader's movement. Wu et al.~\cite{wu2016symmetry} illustrate this problem using two HRP-2 humanoids. For a 54\,kg robot following a leader moving at 0.1\,m/s, their analysis showed that the follower needed to take its next step within approximately 0.72\,s to maintain stability. In simulation, the follower began walking after about 1.3\,s, causing the leader to fall.

Centralized control addresses this problem by coordinating the movements of all robots. Wu et al.~\cite{wu2016symmetry} propose a controller that regulates the object's position and internal forces while commanding all humanoids simultaneously. Their method enables two to four HRP-2 robots to transport objects stably in simulation, but its controller neglects object mass and dynamic effects. Hawley and Suleiman~\cite{hawley2019cooptransport} combine centralized balance control with compliant arms and footstep adjustments. The arms absorb interaction forces, while the footstep planner corrects differences in the robots' relative positions. Their approach allows the distance between robots to change during transport. In physical experiments, two Nao robots carried loads of up to half a robot's mass, although their speed was limited to about 2\,cm/s. Both approaches rely on centralized computation, but neither evaluates the effects of network conditions on cooperative control.

Unlike centralized approaches, leader-follower control can coordinate robot movements through forces transmitted by the shared object, without explicit communication between robots. Wu et al.~\cite{coophumanoidtransport} demonstrate this approach using two physical HRP-2 humanoids. When the leader moves, the follower adjusts its arms in response to forces measured at its wrists, then modifies its footsteps to restore the original hand positions. The robots transported an object while keeping interaction forces generally below 7\,N. However, hand movements continued to oscillate, and the experiments were limited to forward and backward motion.
Gasbarrone et al.~\cite{gasbarrone2024decentralized} extend this approach by allowing the follower to continuously adjust its footstep positions and directions based on interaction forces. This enables more complex movements, including turning, sideways walking, lifting, and obstacle avoidance. However, the leader does not adjust its plan when the follower encounters an obstacle. Learning-based approaches provide another direction. COLA~\cite{du2025cola} demonstrates whole-body collaborative carrying with a human partner, including turning and moving over sloped terrain. However, its effectiveness for cooperation between two humanoids remains untested, as a human partner can compensate for the robot's movements in ways another humanoid may not~\cite{hawley2019cooptransport}. Pandit et al.~\cite{pandit2026decmht} propose decMHT, a decentralized learning-based approach that allows humanoids to carry objects using local observations and shared object-motion commands without direct communication between robots. A policy trained with only two humanoids generalizes to teams of up to ten in simulation and is demonstrated on two physical Digit V3 robots. However, the approach relies on externally provided object-state information, and its performance under network delays remains unexplored.

Physical interaction can guide a follower without explicit communication, but exchanging information helps humanoids synchronize their movements. McGill and Lee~\cite{mcgill2011stretcher} synchronized two stretcher-carrying DARwIn-OP robots through 100\,Hz wireless state updates, achieving a step-phase error of about 5\,ms. However, the robots struggled to maintain their relative distance and fell after an average of 15.5\,s, showing that footstep synchronization alone does not ensure stability. Wen et al.~\cite{dualhumanoidtransport2024} combine scheduled commands with distributed model predictive control to improve leader-follower trajectory tracking. Beyond object carrying, Harmanoid~\cite{liu2025ittakestwo} considers physical interactions between two humanoids, including linked-arm walking and dancing, and incorporates the partner's state and intended body contacts into motion control.

These studies show that physical feedback and synchronized movements alone are not enough for stable collaboration. Forces transmitted through a shared object can help guide a follower, but they may not indicate when a partner is blocked or cannot continue. Similarly, synchronized footsteps do not guarantee that robots maintain their relative positions~\cite{mcgill2011stretcher}. Therefore, humanoids need to exchange timely information about their positions, walking phases, and partner conditions while maintaining balance through local control. However, the effects of network delays and communication failures on these mechanisms remain largely unexplored.

\subsection{Reasoning and Capability Composition}
\label{sec:collab:reasoning}
Humanoids can benefit from collaboration even when they have similar capabilities, as each may hold different information about the environment. CoELA~\cite{coela2023} demonstrates this using two LLM-based agents performing household tasks. Each agent explores independently, maintains knowledge about objects and teammates, and decides when to share information or request help. By coordinating their activities, the two agents delivered 71\% of the target objects, compared with 43\% for a single rule-based agent. Interestingly, removing communication between AI agents had little effect, while removing memory nearly doubled the steps required to complete tasks. These results show how agents can combine their knowledge to work more effectively.

Humanoids may also need to combine different physical capabilities to complete a task. RoCo~\cite{mandi2024roco} uses LLM-based dialogue to help robots understand their teammates' capabilities and decide how to divide the work. For example, one robot holds a dustpan while another sweeps objects into it. In another task, three robotic arms pass cubes between their limited workspaces to reach destinations that no single arm can access directly. The work shows how robots can combine complementary capabilities through shared reasoning, although their motion trajectories are still planned centrally.

Rather than relying on dialogue, humanoids can also collaborate by anticipating each other's actions. COMBO~\cite{combo2025} explores this approach; it uses a VLM to predict what teammates intend to do and a world model to simulate how their combined actions change the environment. Then, based on the predicted outcomes, each robot chooses its next action. The approach improves task success, but repeatedly predicting and evaluating possible outcomes makes planning computationally expensive.

Complementary capabilities can also be combined through learned cooperative policies. HetNet~\cite{seraj2024hetero} explores this approach by training robots with different sensing and acting roles toward a shared goal. In a firefighting scenario, sensing robots search for spreading fires, while other robots use their findings to locate and extinguish them. Although these roles are predefined, HetNet learns how the robots should coordinate their actions. The learned policies can also be transferred to larger teams, reducing the training samples needed to reach comparable performance from about 15 million to 3 million. The authors further demonstrate this cooperation using three physical robots.
However, these approaches do not fully account for how a humanoid's reach, carrying capacity, and available actions change with its posture, balance, and locomotion.

\subsection{Fleet Learning and Cross-Embodiment Transfer}
\label{sec:collab:learning}
Beyond working together to complete tasks, humanoids can also learn from each other's experience to improve their future behavior. Tan~\cite{tan1993iql} explored this idea in a simulated hunting task, where agents shared past experiences or learned policies. Both approaches helped agents learn faster than independent learning, although their final performance was similar. Sharing policies required homogeneous agents, while experience could also be shared between different agents if they could interpret it. This distinction is important for humanoid fleets, whose members may have different physical capabilities, and sharing experiences or policies also introduces communication costs.

Pooling experience from multiple robots can improve learning, but it also induces communication overhead.
OpenBot-Fleet~\cite{openbotfleet2024} enables robots to learn from experiences collected across different homes. It combines images from 72 wheeled robots to train a shared floor-segmentation model, improving perception in unseen environments.
HarvestNet~\cite{chinchali2020cloudCL} reduces data-sharing costs by selecting useful images onboard using model confidence and feature similarity. The cloud uses these images to improve perception models and update sampling thresholds for future collection. This reduces storage, annotation, and computation costs by 66-81\%, but requires target classes to be identified in advance.
Fleet-DAgger~\cite{ifl2023} extends fleet learning by using human interventions to improve a shared policy. It prioritizes robots needing assistance based on failures, uncertainty, and predicted risk, then uses human demonstrations to update the policy for the entire fleet.

Sharing models instead of raw data can reduce communication overhead. Yahya et al.~\cite{yahya2017collective} keep training data on each robot and send only model updates to a central server. Four robotic arms jointly learned to open an unseen door, which a single-robot policy could not achieve. FLEET-MERGE~\cite{wang2023fleetmerge} combines policies trained on different tasks by aligning their internal representations before merging, improving performance by over 50\% compared with baselines. However, performance declines when merging becomes less frequent. Learning While Deploying~\cite{lwd2026}  continuously improves a pretrained VLA policy using experience and human feedback from sixteen dual-arm robots. The updated policy achieved 95\% average success, outperforming imitation fine-tuning on long-horizon tasks. However, each update requires sending the full VLA model to every robot, and the communication overhead is not reported.

When robots have different physical capabilities, sharing knowledge requires aligning their actions, models, or observations. Open X-Embodiment~\cite{oxe2023} combines demonstrations from 22 robot embodiments using a common end-effector action representation. It trains shared RT-X policies using data from nine manipulators, allowing skills learned on one robot to benefit another. In the reported evaluations, RT-1-X improved average task success by 50\% compared with robot-specific methods.
Octo~\cite{ghosh2024octo}  and CrossFormer~\cite{doshi2024crossformer} instead share a transformer backbone while using separate action heads for each robot type. CrossFormer achieved 73\% success compared with 67\% for single-robot training.
Mirage~\cite{chen2024mirage} instead modifies camera observations to let a policy trained on one robot control another without retraining. It outperformed fine-tuned Octo in physical experiments, but requires accurate calibration and is limited to quasi-static tasks involving robots with similar workspaces. These methods show how knowledge can be transferred across different manipulators.

For humanoids, knowledge transfer must account for differences in whole-body structure.
H-Zero~\cite{hzero2025} enables humanoids with different body structures to share locomotion skills. Specifically, it maps the humanoid joints into a common representation and trains a shared policy across simulated robots with varying physical properties. The pretrained policy is then adapted to an unseen humanoid, which achieves stable locomotion after only 30 minutes of fine-tuning. However, its hardware performance initially suffered from drift and jitter.
Its results also show that adding embodiment descriptors or upper-body joints can reduce transfer performance.
XHugWBC~\cite{crosshumanoidwbc2026} further reduces the need for robot-specific training. It trains a shared whole-body controller using simulated humanoids with different joint structures and physical properties. The resulting policy controlled seven physical humanoids without fine-tuning and achieved about 85\% of the reward of robot-specific controllers in simulation. However, it requires a common command interface, and some larger humanoids showed greater performance degradation when transferred to hardware.
Cross-embodiment learning also extends to fall recovery. Spraggett~\cite{getupmorph2025} trains a shared recovery policy across small simulated humanoids without providing explicit information about their body structures. These studies demonstrate the benefits of knowledge sharing, but rarely examine how it works across deployed humanoid fleets. Communication costs and delays can slow the exchange of experiences and model updates, while unsafe or compromised updates may spread across robots~\cite{fedpoison2024,propagunsafe2026}. How these factors affect the reliability and safety of fleet learning remains largely unexplored.

\begin{table*}[t]
\centering
\caption{Experience and knowledge sharing mechanisms compared by what is shared, the compatibility required, and embodiment evidence. Limited works measure the communication cost of sharing.}
\label{tab:knowledge}
\scriptsize
\setlength{\tabcolsep}{3pt}
\renewcommand{\arraystretch}{1.05}
\resizebox{\textwidth}{!}{%
\begin{tabular}{>{\raggedright\arraybackslash}p{2.8cm} >{\raggedright\arraybackslash}p{4.2cm} >{\raggedright\arraybackslash}p{5.0cm} >{\raggedright\arraybackslash}p{1.6cm}}
\toprule
\textbf{Work} & \textbf{What is shared} & \textbf{Compatibility required} & \textbf{Evidence} \\
\midrule
OpenBot-Fleet~\cite{openbotfleet2024} & Raw driving frames & Same robot design & RR \\
HarvestNet~\cite{chinchali2020cloudCL} & Rare, informative samples & Same perception task & RR \\
Fleet-DAgger~\cite{ifl2023} & Human interventions & Common state and action spaces & HS, RS, RR \\
Collective GPS~\cite{yahya2017collective} & Model updates to a shared backbone & Shared backbone, robot-specific heads & RR \\
FLEET-MERGE~\cite{wang2023fleetmerge} & Trained policy parameters & Same architecture; unit alignment & RS \\
LWD~\cite{lwd2026} & Rollouts, interventions, redeployed policy & Common pretrained policy & RR \\
Mirage~\cite{chen2024mirage} & URDF and renderer; no runtime exchange & Similar grippers and workspace & RR \\
Open X-Embodiment~\cite{oxe2023} & Trajectories in a canonical interface & End-effector actions, canonical view & RR \\
H-Zero~\cite{hzero2025} & Pretrained cross-humanoid policy & Shared joint space, aligned directions & HS \\
XHugWBC~\cite{crosshumanoidwbc2026} & One whole-body policy & Aligned observations and actions & HS, HR \\
\bottomrule
\end{tabular}
}
\vspace{-0.2in}
\end{table*}

\subsection{Summary and Lessons Learned}
\label{sec:collab:summary}

These forms of collaboration differ in what humanoids combine. In physically coupled execution, humanoids combine forces and must exchange information within the control loop. In reasoning and capability composition, they combine knowledge and skills at the planning timescale, while in fleet learning they share experience. Reasoning and capability composition are tested only with simulated agents or robotic arms, and experience sharing with wheeled robots. Three lessons follow.

\textbf{Physical coupling and inference can reduce communication, but cannot replace it.} A follower can infer the leader's intent from forces transmitted through the shared object~\cite{gasbarrone2024decentralized,du2025cola}, and COMBO predicts its partners' actions without exchanging messages~\cite{combo2025}. However, forces cannot reveal a blocked partner or gradual drift~\cite{gasbarrone2024decentralized,mcgill2011stretcher}, and COMBO's predictions are slow and error-prone. Communication is most valuable for information a robot cannot observe or infer by itself: HetNet's sensorless robots cannot act without it~\cite{seraj2024hetero}.

\textbf{All forms lack a usable description of the contributing body.} Harmanoid's partner state lifts success from at most 20\% to 92\% but is read from the simulator~\cite{liu2025ittakestwo}, GPT-4 misjudges its arm's reachability about a third of the time~\cite{mandi2024roco}, and embodiment descriptors can even reduce cross-humanoid transfer~\cite{hzero2025}. Therefore, the motion envelope proposed for coordination should be extended from where a humanoid will be to what it can safely contribute in its current state and how its skills map to other bodies.

\textbf{Studies report the benefits of collaboration but rarely its costs.} Among the physical collaboration studies, only one describes the network it uses~\cite{mcgill2011stretcher}. None of the works in Table~\ref{tab:knowledge} measures communication cost or examines how unsafe updates could spread~\cite{fedpoison2024,propagunsafe2026}, even though LWD sends a full VLA model to every robot after each update~\cite{lwd2026}. Therefore, it is unclear when a fleet should collaborate by exchanging messages, inferring partner behavior, or sharing models.

\section{Social Intelligence and Human Interaction}
\label{sec:social}

Humans cannot be treated as merely additional teammates: their intentions are not directly observable, their expectations are not defined by the system, and they hold responsibility and authority over the fleet. Therefore, networked humanoids must infer what humans intend, communicate their own intentions clearly, adapt to human behavior, and preserve meaningful human control. This section reviews these requirements and examines how they change when multiple humanoids interact with and are supervised by the same people.

\textbf{Why social interaction is different for humanoids.} Humanoid embodiment creates several challenges for human interaction. First, a human-like body can influence trust:~\cite{roesler2024trustdynamics} found that participants trusted a robotic arm more than a humanoid robot. Second, people interpret a robot's motion as a signal of its goal even when that motion primarily serves a physical task~\cite{dragan2013legibility}. Third, humanoids operate close to people and may make physical contact, making perceived safety important in addition to physical safety~\cite{gentlehumanoid2025,rubagotti2022perceived}. These challenges extend from individual interaction to the fleet level. When several humanoids work with the same person, each may maintain its own estimate of that person's state, and these estimates must remain consistent across the fleet~\cite{simtom_hrteams2025}. 

\subsection{Intentions: Social State Inference and Common Ground}
\label{sec:social:intent}
Effective interaction requires a humanoid to infer what a person intends, what the person believes, and what information they share. Intent inference is the most mature of these problems.
Hoffman et al.~\cite{hoffman2024intent} review two main approaches: probabilistic models that estimate human intentions from observed actions, and neural models that learn these relationships from data. Probabilistic models depend on assumptions about human behavior, while neural models require sufficient training data.
The study~\cite{bayesianintent_hrc2024} shows how a robot can use inferred human intentions to adapt its actions. They combine head orientation, hand orientation, and hand movement in a Bayesian network to predict which object a person intends to reach for. This improves prediction accuracy by 85\% over the best single-cue baseline, with an average prediction time of 2.69~ms. A UR5 robot then uses the predicted target to adjust its task plan and avoid interfering with the person's movement.
Predicting a person's intentions is not enough when collaboration also depends on what the person knows about the environment. The work~\cite{beliefstate_hrteam2025} addresses this by tracking objects in a scene graph and estimating the person's knowledge based on what they could observe. Their estimates agreed with participants' answers about 55-75\% of the time.
Nicolescu et al.~\cite{simtom_hrteams2025} enable a Baxter and a PR2 to infer a person's current task from hand movements and exchange task-status messages to avoid duplicating work. Verbal dialogue resolves conflicts between the person and robots. However, the framework does not address conflicting estimates of human intentions across robots.

Understanding a person's intentions is only part of effective interaction. Humans and robots must also establish a shared understanding of the task and environment. Anikina et al.~\cite{anikina2025ground} explain how this common ground develops through dialogue, clarification, and nonverbal cues. For robot teams, Scheutz et al.~\cite{scheutz2017shared} propose shared mental models that represent teammates' beliefs, goals, roles, and task progress. Their framework allows corrections received by one robot to update the models of others, although it has only been partially implemented.

Humanoid robots can establish common ground by directing attention to objects in their surroundings. Neuber et al.~\cite{neuber2026gaze} review gaze-following systems that identify what a person is looking at, but find that many rely on controlled environments and rarely report response times. Conversely, robots can use gestures to help people identify objects. Sauppé and Mutlu~\cite{sauppe2014deictics} show that touching or holding up an object is particularly effective. Common ground can also improve through repeated interaction. Bärmann et al.~\cite{barmann2024incremental} enable ARMAR-6 to learn from human corrections and reuse them for similar requests, reducing the need for further corrections. These findings highlight a challenge for networked humanoids: how to share and maintain a consistent understanding across robots without spreading incorrect assumptions.

\subsection{Expectations: Legibility, Proximity, Norms, and Trust}
\label{sec:social:expect}
Preventing physical harm does not necessarily make people feel safe around humanoids. Rubagotti et al.~\cite{rubagotti2022perceived} show that perceived safety also depends on the robot's appearance, movements, and proximity to people. Akalin et al.~\cite{akalin2022safe} further find that perceived safety is closely related to comfort, sense of control, and trust. In contrast, recent control methods mainly focus on physical safety. GentleHumanoid~\cite{gentlehumanoid2025} enables a Unitree G1 to adapt its upper-body compliance during hugging and sit-to-stand assistance, using adjustable force thresholds of 5-15~N. Busellato et al.~\cite{busellato2026uapcbf} adjust a robotic arm's safety margin based on uncertainty in predicted human hand movements, reducing safety-zone violations by about tenfold. However, neither study measures how safe people feel during interaction, leaving perceived safety unclear. Safe motion does not necessarily make a robot's intentions clear. Dragan et al.~\cite{dragan2013legibility} distinguish predictable motion, which follows the path people expect, from legible motion, which helps people recognize the robot's goal early. In a collaborative task with HERB, legible motion reduced the time needed to infer the robot's goal by 33\%, but did not improve perceived safety~\cite{dragan2015motion}. These findings show that making robot movements easier to understand does not necessarily make people feel safer.

Beyond making its intentions clear, a humanoid must also respect people's expectations of personal space. Reviewing 39 studies, Nahum et al.~\cite{nahum2026proxemics} show that comfortable interaction distances depend on human movement and context. However, most studies were conducted in controlled laboratories, and none examined cultural differences. Lawrence et al.~\cite{lawrence2025norms} find that violating social norms can reduce comfort, trust, and acceptance, while Saettone et al.~\cite{saettone2026culture} highlight how these expectations vary across cultures. To support socially aware navigation, Mu et al.~\cite{mu2026social} track human group movements and guide a Roban humanoid toward potential interaction partners while avoiding other groups. However, their experiments rely on motion capture and evaluate navigation performance rather than people's comfort.

Respecting social norms can help build trust, but trust should also reflect how reliably a robot performs. Mehrotra et al.~\cite{mehrotra2024appropriatetrust} review methods for aligning human trust with system reliability and find that explanations and human-like appearance do not consistently improve this alignment. Measuring trust is also challenging. Roesler et al.~\cite{roesler2024trustdynamics} found that robot failures reduced reported trust without corresponding changes in participants' speech behavior. Campagna and Rehm~\cite{campagna2025trustassessment} further show that trust assessment relies mainly on questionnaires, while sensor-based methods lack standardized approaches. Trust also works in the opposite direction. Vinanzi et al.~\cite{collabmind_intent2021} enable a simulated Sawyer robot to estimate a human partner's reliability and take control when the partner repeatedly fails. However, how trust develops and changes when people interact with multiple humanoids remains largely unexplored. 

\vspace{-0.1in}
\subsection{Responsibilities and Control: Teaming, Teleoperation, and Oversight}
\label{sec:social:resp}
Human-humanoid teaming requires deciding not only who performs each task, but also when robots should guide people or follow their preferences. Nikolaidis et al.~\cite{nikolaidis2017mutual} enable an assistive robot arm to estimate how willing a user is to change their goal. The robot guides adaptable users toward a better strategy while following the preferences of those unwilling to change. In an experiment with 51 participants, this approach improved task performance while maintaining high user trust. Lagomarsino et al.~\cite{lagomarsino2026role} review methods that adapt human-robot roles based on fatigue, cognitive load, and preferences. However, most approaches remain task-specific, with limited validation in real-world settings.

Adapting roles also requires understanding how much responsibility people can manage. Wong and Seet~\cite{wong2017workload} show that supervising more robots increases workload and can reduce situation awareness, while greater automation may encourage complacency. To support supervision, Ali et al.~\cite{ali2025situation} estimate whether an operator has high or low situation awareness using eye tracking and task behavior, achieving 74\% accuracy in a simulated two-robot task. Beyond individual workload, team relationships also matter. You and Robert~\cite{you2023subgroup} show that strong attachment to individual robots can divide human-robot teams into smaller groups, while a shared team identity may help prevent this division. These findings suggest that adapting human-robot responsibilities should also preserve team cohesion.

Human control of humanoids ranges from direct teleoperation to high-level supervision. In recent teleoperation systems, people guide the robot's movements while onboard policies maintain balance. H2O~\cite{he2024h2o} captures human motion through an RGB camera and trains its policy on movements the humanoid can perform. OmniH2O~\cite{he2024omnih2o} allows the same control policy to receive commands from VR, cameras, or GPT-4o through a common pose interface. HOMIE~\cite{homie2025} takes a different approach, using foot pedals, an arm exoskeleton, and sensing gloves for more precise control. In experiments, it nearly halved task completion time compared with VR-based teleoperation.

As humanoids become more autonomous, people must decide when to intervene rather than control every movement. Verhagen et al.~\cite{verhagen2024meaningful} emphasize that meaningful human control requires operators to understand robot behavior and remain able to intervene. For multiple robots, Fleet-DAgger~\cite{ifl2023} directs limited human attention to robots that have failed, are uncertain, or face higher risks, while using interventions to improve a shared policy. However, it assumes supervisors can switch between robots immediately and does not consider network delays. How to ensure timely human intervention across networked humanoid fleets therefore remains unclear.
\vspace{-0.1in}

\subsection{Summary and Lessons Learned}
\label{sec:social:summary}

Few studies examine social interaction involving multiple humanoids. Research on intention inference, legible motion, and mutual adaptation mainly uses robotic arms or simulations~\cite{bayesianintent_hrc2024,beliefstate_hrteam2025,dragan2015motion,chan2025argaze,nikolaidis2017mutual}. Studies with physical humanoids instead focus on gestures, physical contact, navigation, and teleoperation~\cite{sauppe2014deictics,barmann2024incremental,gentlehumanoid2025,mu2026social,he2024omnih2o,homie2025}. Most involve a single robot, with limited evidence from multi-robot settings~\cite{simtom_hrteams2025,ifl2023}. Three lessons follow.

\textbf{Social state stays local to the robot that estimated it.}
Intent, beliefs, and user corrections are estimated in one interaction and kept by one robot. When two robots track the same person, how they reconcile disagreeing estimates is left unspecified~\cite{simtom_hrteams2025}, and ARMAR-6 reuses stored corrections without verifying them~\cite{barmann2024incremental}. However, a worker expects consistent behavior from every humanoid, such as the same passing distance. Shared social state should carry its source, scope, uncertainty, and validity period, as coordination and body-centric state should, so that sharing does not spread errors along with knowledge.

\textbf{Human authority and attention must be managed at fleet scale.} Current work follows a control structure suited to networked humanoids, where humans provide motion goals while onboard policies control balance~\cite{he2024h2o,he2024omnih2o,homie2025}. However, supervising multiple robots is more challenging. As one operator oversees more robots, workload increases and situation awareness declines~\cite{wong2017workload}. This can lead to missed interventions and a loss of meaningful human control~\cite{verhagen2024meaningful}. Fleet-DAgger addresses this challenge by directing operator attention to robots with higher risks or uncertainty~\cite{ifl2023}, but assumes identical robots and does not account for network latency. Changes in roles or control authority must also reach every affected robot and remain traceable to the person who approved them.

\section{Evaluation, Safety, and Open Problems in Networked Humanoid Intelligence}
\label{sec:evaluation}

This section examines how networked humanoid intelligence should be evaluated, what new safety risks arise when humanoids operate as a connected population, and which problems remain open. We first review the available evaluation resources and identify the gaps that make results difficult to compare across studies. We then distinguish risks that arise within an individual humanoid from those introduced or amplified by communication and interaction across the fleet. Finally, we use the limitations identified throughout this survey to outline directions for future research.

\subsection{Platforms, Simulators, Datasets, and Benchmarks}
\label{sec:eval:resources}

Evaluation resources address individual components of networked humanoid intelligence, but none captures them together. Based on the requirements identified in the previous sections, we compare these resources along five dimensions: humanoid embodiment, multiple agents, network conditions, physical interaction between agents, and human presence in Table~\ref{tab:resources}. Humanoid hardware is becoming increasingly accessible, ranging from low-cost research platforms for learning-based control~\cite{berkeleyhumanoid2024} to commercial systems such as the Unitree H1 and G1~\cite{sheng2025smartbot,berkeleyhumanoid2024}. However, physical experiments with multiple humanoids still rely largely on small platforms such as NAO and DARwIn-OP~\cite{coltin2010multi,mcgill2011stretcher,hawley2019cooptransport}. GPU-parallel simulators such as Isaac Gym and Isaac Lab~\cite{makoviychuk2021isaacgym,mittal2025isaaclab}, together with frameworks such as Humanoid-Gym~\cite{humanoidgym2024} and benchmarks such as HumanoidBench~\cite{sferrazza2024humanoidbench}, simulate whole-body and contact dynamics; Isaac Lab additionally supports multi-agent environments and configurable actuation delays. However, these resources are typically used to train and evaluate a single humanoid, and none models communication between agents over a realistic network. Datasets such as Humanoid Everyday~\cite{humanoideveryday2025}, which includes human--humanoid interaction tasks, human-motion and humanoid-retargeted motion corpora~\cite{mahmood2019amass,phuma2025}, and large-scale manipulation datasets that are either cross-embodiment~\cite{oxe2023} or collected on a single platform across many scenes~\cite{khazatsky2024droid} support learning and transfer, but capture neither interaction among multiple robots nor network conditions. Multi-agent embodied benchmarks~\cite{chen2025emos,mandi2024roco,coela2023}, surveyed in~\cite{embodiedmasurvey2025}, provide team settings but rarely include humanoids and typically measure communication cost in tokens, model queries, or environment steps rather than network delay or traffic. Robot soccer remains a long-running testbed for multi-humanoid systems, both on physical robots~\cite{coltin2010multi} and in simulation~\cite{nadiri2025swarm}. Conversely, network testbeds such as O-RACES~\cite{wagner2026races} connect simulated humanoids through real Open RAN infrastructure, but involve neither physical interaction among the robots nor human participants. None of the reviewed resources combines all five dimensions needed to evaluate networked humanoid intelligence as an integrated system.

\begin{table*}[t]
\centering
\caption{Evaluation resources assessed against the five properties that networked humanoid intelligence jointly requires. \yes: provided; \partc: partial or optional; \no: absent.}
\label{tab:resources}
\scriptsize
\setlength{\tabcolsep}{3pt}
\renewcommand{\arraystretch}{1.05}
\resizebox{\textwidth}{!}{%
\begin{tabular}{>{\raggedright\arraybackslash}p{3.6cm} >{\raggedright\arraybackslash}p{1.9cm} c c c c c >{\raggedright\arraybackslash}p{3.4cm}}
\toprule
\textbf{Resource} & \textbf{Type} & \textbf{Humanoid} & \makecell{\textbf{Multiple}\\\textbf{agents}} & \makecell{\textbf{Network}\\\textbf{modeled}} & \makecell{\textbf{Physical}\\\textbf{interaction}} & \textbf{Humans} & \textbf{Main limitation} \\
\midrule
Isaac Gym / Isaac Lab~\cite{makoviychuk2021isaacgym,mittal2025isaaclab} & Simulator & \yes & \partc & \no & \partc & \no & Ideal inter-agent channel \\
Humanoid-Gym, HumanoidBench~\cite{humanoidgym2024,sferrazza2024humanoidbench} & Framework, benchmark & \yes & \no & \no & \no & \no & Single humanoid \\
Humanoid Everyday~\cite{humanoideveryday2025} & Dataset & \yes & \no & \no & \no & \partc & No robot-robot interaction\\
Open X-Embodiment, DROID~\cite{oxe2023,khazatsky2024droid} & Dataset & \no & \no & \no & \no & \no & Arms; no whole-body state \\
Habitat-MAS~\cite{chen2025emos} & Benchmark & \no & \yes & \no & \no & \no & Cost counted in tokens \\
RoCoBench~\cite{mandi2024roco} & Benchmark & \partc & \yes & \no & \partc & \partc & Humanoid as a fixed-base arm \\
TDW-MAT, C-WAH~\cite{coela2023} & Benchmark & \no & \yes & \partc & \no & \partc & Messages cost environment steps \\
RoboCup SPL, Webots soccer~\cite{coltin2010multi,nadiri2025swarm} & Testbed & \yes & \yes & \partc & \partc & \no & Adversarial game \\
O-RACES testbed~\cite{wagner2026races} & Network testbed & \partc & \yes & \yes & \no & \no & Simulated humanoids \\
Harmanoid setup~\cite{liu2025ittakestwo} & Simulation & \yes & \yes & \no & \yes & \no & No network, no humans \\
Simulation theory of mind~\cite{simtom_hrteams2025} & HRI testbed & \partc & \yes & \no & \no & \yes & Network not characterized \\
Interactive fleet learning~\cite{ifl2023} & Benchmark & \partc & \yes & \no & \no & \yes & Humans as supervisors only \\
\bottomrule
\end{tabular}
}
\vspace{-0.2in}
\end{table*}

\subsection{What Current Evaluations Cannot Compare}
\label{sec:eval:gaps}

Current evaluations are difficult to compare because they differ along five fundamental dimensions.

\emph{Embodiment:} Few reviewed mechanisms are evaluated on physical or simulated humanoids, and none of the communication mechanisms is tested on a physical humanoid. Physical-humanoid evaluations are concentrated mainly in one-to-one social interaction and cooperative carrying by pairs of small robots.

\emph{Cost units:} Communication cost is reported in incompatible units, including bits, tokens, environment steps, dialogue turns, and solver rounds. These measures do not reveal the actual network resources required. For example, the reported results cannot determine whether COMBO's additional model calls impose a greater system cost than CoELA's explicit messages~\cite{combo2025,coela2023}.

\emph{Networking:} Coordination and collaboration studies typically assume reliable communication, whereas networking studies use synthetic traffic or traffic from isolated perception and control tasks. As a result, the effect of realistic network impairments on team behavior remains unclear.

\emph{Physical outcomes:} Network performance is rarely connected to physical consequences. Only two reviewed studies report humanoid falls under communication delay, both with simulated humanoids:~\cite{zhu2020remotewbc,wagner2026races}, and none measures interaction forces under network impairment.

\emph{Human outcomes:} Human interaction is evaluated mainly in one-human-one-robot settings using questionnaires and behavioral measures. Fleet-level outcomes, such as how trust changes across robots or how supervision workload scales with fleet size, remain largely unmeasured.

These gaps suggest the need for consistent reporting rather than a single benchmark. Evaluations should state the embodiment level, report communication volume and timing requirements alongside task-specific cost measures, and characterize performance under varying delay, jitter, and packet loss. For physically coupled tasks, they should additionally report outcomes such as falls, balance-margin violations, and peak interaction forces. Fleet evaluations should show how both communication cost and task performance scale with team size, and how long exchanged information remains useful. When humans are involved, evaluations should also measure intervention frequency, operator situation awareness, and whether reported trust is consistent with the robot's actual reliability.

\subsection{Safety: The Per-Robot Floor and the Risks the Network Adds}
\label{sec:eval:safety}

Networking does not change the basic mechanisms that keep an individual humanoid safe, but it introduces risks that cannot be managed by one robot alone. We define the \emph{per-robot safety floor} as the protection that remains available without network support, including onboard stabilization, bounded interaction forces~\cite{gentlehumanoid2025}, control-barrier-function safety filters or policies trained with them~\cite{busellato2026uapcbf,cbfrl2025}, and protective falling~\cite{safefall2025}. Every network-dependent function should fall back to this local protection when communication fails. Zhu et al.~\cite{zhu2020remotewbc}, for example, keep fast feedback onboard and reuse the most recent edge result when a new one arrives too late.

Above this floor, networking introduces three additional risks. \emph{Staleness} occurs when delayed information no longer represents the current physical or team state. Delay and packet loss can degrade control stability~\cite{4118465,9380899}, reactive medium access can cause simulated humanoid falls~\cite{wagner2026races}, and a stumble can invalidate a robot's previous task and motion commitments. \emph{Correlation} occurs when robots depend on the same communication or coordination process. Synchronized transmissions, for example, can make shared information stale for several robots at the same time~\cite{zhao2026desync}.

\emph{Propagation} occurs when information shared across the fleet carries errors or malicious behavior from one robot to others. Poisoned updates can corrupt a shared model; FedREDefense~\cite{yueqi2024fedredefense} addresses this problem by testing whether an update can be reconstructed from distilled local knowledge, which helps distinguish genuine updates from crafted attacks~\cite{fedpoison2024}. In LLM-controlled teams, compromising one robot can also spread unsafe actions through inter-robot messages~\cite{propagunsafe2026,embodiedaisafety2026}, while self-interested agents can learn to send adversarial messages~\cite{zhou2021review}. Similar propagation can occur in human interaction. Trust developed with one robot may extend to other members of the fleet, complicating trust calibration~\cite{yu2026ioh}, while approvals made without sufficient understanding can weaken meaningful human control~\cite{verhagen2024meaningful}. Networked humanoid safety therefore requires containment in addition to per-robot protection: information should be checked for freshness before entering control loops, every network-dependent function should have a local fallback, shared knowledge should be verified before fleet-wide deployment, and the origin of commitments, social estimates, and human approvals should remain traceable.
\vspace{-0.15in}
\subsection{Open Problems and Research Agenda}
\label{sec:eval:open}

The limitations identified throughout this survey point to six directions for future research.

\textbf{Control-aware networking.} Communication is still evaluated mainly through network metrics, with limited reviewed studies relating network performance to humanoid falls. Yet the consequences of delay depend on the robot's physical state; walking, for example, is more delay-sensitive than standing~\cite{zhu2020remotewbc}. Future systems should therefore allow controllers to expose gait- and contact-dependent communication deadlines to network schedulers such as O-RACES~\cite{wagner2026races}. Evaluation should then connect delay, jitter, and packet loss directly to physical outcomes such as falls and balance violations under realistic traffic.

\textbf{Compact body-centric messages.} Networked humanoids need to exchange contact state, gait phase, footstep intent, and predicted whole-body occupancy, but compact representations for these quantities remain limited. One direction is the motion envelope, which could represent predicted body occupancy, planned footsteps, allowable motion deviation, and how long this information remains useful. For physically coupled robots, this representation could be complemented by compact contact and gait-phase messages. Such representations would allow a robot to determine whether a teammate's plan remains physically compatible with its own motion without exchanging complete trajectories.

\textbf{Physically grounded coordination.} Current allocation methods represent capability mainly through utilities, symbolic skills, or geometric constraints, while largely ignoring how balance, posture, and load affect whether a humanoid can safely execute an assignment. These constraints also change during execution. Propose-and-validate architectures~\cite{chen2024scalablellm,liu2025coherent} provide a useful starting point: before accepting an assignment, each humanoid could query its local whole-body controller to verify that the task remains feasible under its current physical state.

\textbf{Network- and embodiment-aware knowledge sharing.} Fleet learning largely treats communication as an external assumption, while cross-humanoid transfer remains focused mainly on locomotion and existing pipelines are not evaluated under corrupted updates. Future systems should therefore treat knowledge sharing as a joint learning, embodiment, and networking problem. Initial steps include measuring update size and rollout latency in systems such as LWD~\cite{lwd2026}, separating transferable model components from robot-specific adaptations~\cite{hzero2025}, and deploying updates gradually with mechanisms for detecting harmful updates, such as reconstruction-based filtering~\cite{yueqi2024fedredefense}. The goal is to improve heterogeneous humanoids across the fleet while keeping communication cost measurable and limiting how far an incorrect or malicious update can propagate.

\textbf{Fleet-level social consistency.} Social information learned by one humanoid is rarely reconciled with estimates held by other robots, and changes in human roles or preferences are not systematically propagated across the fleet. A shared social-state representation could address this gap by associating each estimate with its source, scope, uncertainty, and the period for which it remains useful. Such a representation would extend the task-state exchange of Nicolescu et al.~\cite{simtom_hrteams2025} from task progress to human-related information. Evaluation should then measure whether different humanoids behave consistently toward the same person while still adapting to changes in the interaction context.

\textbf{Accountable oversight.} Human attention becomes a limited resource as the number of supervised robots increases. Approaches such as Fleet-DAgger~\cite{ifl2023} allocate this attention according to learning needs, but humanoid fleets must also consider the physical consequences of delayed intervention. Future systems should therefore allocate operator attention according to both task needs and physical risk, while coordinating human intervention with available network resources. Recording each approval together with the information available to the operator would further make decisions traceable and clarify when and why human authority was exercised.

Across all four functions, these directions point to a common requirement: information freshness, embodiment, and human accountability must be considered jointly, rather than communication, coordination, collaboration, and social interaction being treated as independent problems.

\section{Conclusion}
\label{sec:conclusion}
Humanoid robots are moving from standalone systems toward connected fleets, where robots exchange information, coordinate actions, work together, and interact with humans. This survey reviewed networked humanoid intelligence through four key functions: communication and networking, coordination, collaboration, and social intelligence and human interaction. We examined how these functions support fleet operation and how the physical characteristics of humanoids introduce new challenges. In particular, balance, whole-body motion, physical contact, and hardware differences place additional demands on information exchange and joint decision-making. Despite recent progress, most studies treat these functions in isolation and pay little attention to network conditions, differences between humanoids, or safety across the fleet. Future work should jointly consider communication, planning, physical control, and human oversight, so that humanoid fleets can work safely alongside people, adapt to changing environments, and improve by sharing knowledge and experience.

\bibliographystyle{ACM-Reference-Format}
\bibliography{mybib}

@article{akalin2022safe,
title = {Do you feel safe with your robot? Factors influencing perceived safety in human-robot interaction based on subjective and objective measures},
journal = {International Journal of Human-Computer Studies},
volume = {158},
pages = {},
year = {2022},
issn = {1071-5819},
author = {Neziha Akalin and others}
}

@INPROCEEDINGS{ali2025situation,
  author={Ali, Arsha and Robert, Lionel P. and Tilbury, Dawn M.},
  booktitle={Proc. IEEE International Conference on Robot and Human Interactive Communication (RO-MAN)}, 
  title={Estimating Situation Awareness for Human-Robot Teaming}, 
  year={2025},
  volume={},
  number={},
  pages={1216-1221}}

@article{humanoidworldmodels2025,
  title={Humanoid world models: Open world foundation models for humanoid robotics},
  author={Ali, Muhammad Qasim and Sridhar, Aditya and Matiana, Shahbuland and Wong, Alex and Al-Sharman, Mohammad},
  journal={Preprint arXiv:2506.01182},
  year={2025}
}

@article{alonso2019distributed,
  title={Distributed multi-robot formation control in dynamic environments},
  author={Alonso-Mora, Javier and Montijano, Eduardo and N{\"a}geli, Tobias and Hilliges, Otmar and Schwager, Mac and Rus, Daniela},
  journal={Autonomous Robots},
  volume={43},
  number={5},
  pages={1079--1100},
  year={2019},
  publisher={Springer}
}

@INPROCEEDINGS{andrei2024digital,
  author={Andrei, Vlad C. and Li, Xinyang and Fees, Maresa and Feik, Andreas and others},
  booktitle={Proc. IEEE International Symposium on Joint Communications \& Sensing}, 
  title={A Digital Twinning Platform for Integrated Sensing, Communications and Robotics}, 
  year={2024},
  volume={},
  number={},
  pages={}}

@inproceedings{anikina2025ground,
    title = "Building Common Ground in Dialogue: A Survey",
    author = "Anikina, Tatiana  and
      Leippert, Alina  and
      Ostermann, Simon",
    booktitle = "Proceedings of the 2nd LUHME Workshop",
    month = oct,
    year = "2025",
    address = "Bologna, Italy",
    pages = "3--28",
}

@article{barmann2024incremental,
  title={Incremental learning of humanoid robot behavior from natural interaction and large language models},
  author={B{\"a}rmann, Leonard and Kartmann, Rainer and Peller-Konrad, Fabian and Niehues, Jan and others},
  journal={Frontiers in Robotics and AI},
  volume={11},
  pages={1455375},
  year={2024},
  publisher={Frontiers Media SA}
}

@INPROCEEDINGS{homie2025, 
    AUTHOR    = {Qingwei Ben AND Feiyu Jia and others}, 
    TITLE     = {{HOMIE: Humanoid Loco-Manipulation with Isomorphic Exoskeleton Cockpit}}, 
    BOOKTITLE = {Proceedings of Robotics: Science and Systems}, 
    YEAR      = {2025}, 
    ADDRESS   = {LosAngeles, CA, USA}, 
    MONTH     = {June},
}

@ARTICLE{connectedrobotics2026,
  author={Bezerra, Ranulfo and Tadokoro, Satoshi and Ohno, Kazunori},
  journal={IEEE Internet of Things Journal}, 
  title={AI-IoT-Robotics Integration: Survey of Frameworks, Emerging Trends, and the Path Toward Connected Robotics}, 
  year={2026},
  volume={13},
  number={10},
  pages={20398-20412}}

@article{bin2025visualperception,
title = {A survey on the visual perception of humanoid robot},
journal = {Biomimetic Intelligence and Robotics},
volume = {5},
number = {1},
pages = {100197},
year = {2025},
issn = {2667-3797},
author = {Teng Bin and Hanming Yan and Ning Wang and Milutin N. Nikolić and Jianming Yao and Tianwei Zhang},
}

@article{groot2025,
  title={Gr00t n1: An open foundation model for generalist humanoid robots},
  author={Bjorck, Johan and Casta{\~n}eda, Fernando and Cherniadev, Nikita and Da, Xingye and Ding, Runyu and Fan, Linxi and Fang, Yu and Fox, Dieter and others},
  journal={Preprint arXiv:2503.14734},
  year={2025}
}

@article{black2024pi0,
  title={$\pi_{0}$: A Vision-Language-Action Flow Model for General Robot Control},
  author={Black, Kevin and Brown, Noah and Driess, Danny and Esmail, Adnan and Equi, Michael and Finn, Chelsea and Fusai, Niccolo and Groom, Lachy and Hausman, Karol and others},
  journal={RSS},
  year={2025}
}

@article{busellato2026uapcbf,
title = {Uncertainty Aware-Predictive Control Barrier Functions: Safer human–robot interaction through probabilistic motion forecasting},
journal = {Robot. Auton. Syst.},
volume = {197},
pages = {105291},
year = {2026},
issn = {0921-8890},
author = {Lorenzo Busellato and Federico Cunico and others},
}

@article{campagna2025trustassessment,
author = {Campagna, Giulio and Rehm, Matthias},
title = {A Systematic Review of Trust Assessments in Human–Robot Interaction},
year = {2025},
issue_date = {June 2025},
publisher = {Association for Computing Machinery},
address = {New York, NY, USA},
volume = {14},
number = {2},
journal = {ACM Transactions on J. Hum.-Robot Interact.},
month = jan,
articleno = {30},
numpages = {35}
}

@article{cao2024humanoid,
author = {Cao, Longbing},
title = {Humanoid Robots and Humanoid AI: Review, Perspectives and Directions},
year = {2025},
issue_date = {March 2026},
publisher = {ACM},
address = {New York, NY, USA},
volume = {58},
number = {4},
issn = {0360-0300},
journal = {ACM Comput. Surv.},
month = oct,
articleno = {97},
numpages = {37}
}

@article{chan2025argaze,
  title={Improving human--robot collaboration through augmented reality and eye gaze},
  author={Chan, Wesley P and Crouch, Morgan and Hoang, Khoa and Chen, Charlie and others},
  journal={ACM Trans. Hum.-Robot Interact.},
  volume={14},
  number={3},
  pages={1--19},
  year={2025},
  publisher={ACM New York, NY}
}

@inproceedings{chen2025emos,
 author = {Chen, Junting and Yu, Checheng and Zhou, Xunzhe and Xu, Tianqi and Mu, Yao and Hu, Mengkang and Shao, Wenqi and Wang, Yikai and others},
 booktitle = {Proc. ICLR},
 pages = {},
 title = {EMOS: Embodiment-aware Heterogeneous Multi-robot Operating System with LLM Agents},
 year = {2025}
}

@inproceedings{chen2024mirage,
    title={Mirage: Cross-Embodiment Zero-Shot Policy Transfer with Cross-Painting},
    author = {Lawrence Yunliang Chen and Kush Hari and Karthik Dharmarajan and Chenfeng Xu and others},
    booktitle = {Proceedings of Robotics: Science and Systems},
    address  = {Delft, Netherlands},
    year = {2024},
    }

@article{autonomousmrnav2024,
title = {A survey of autonomous robots and multi-robot navigation: Perception, planning and collaboration},
journal = {Biomimetic Intelligence and Robotics},
volume = {5},
number = {2},
pages = {100203},
year = {2025},
issn = {2667-3797},
author = {Weinan Chen and Wenzheng Chi and Sehua Ji and Hanjing Ye and Jie Liu and others}
}

@INPROCEEDINGS{chen2024scalablellm,
  author={Chen, Yongchao and Arkin, Jacob and Zhang, Yang and Roy, Nicholas and Fan, Chuchu},
  booktitle={Proc. IEEE ICRA}, 
  title={Scalable Multi-Robot Collaboration with Large Language Models: Centralized or Decentralized Systems?}, 
  year={2024},
  volume={},
  number={},
  pages={4311-4317}}

@InProceedings{chinchali2020cloudCL,
author="Chinchali, Sandeep and Pergament, Evgenya and Nakanoya, Manabu and others",
title="Sampling Training Data for Continual Learning Between Robots and the Cloud",
booktitle="Experimental Robotics",
year="2021",
publisher="Springer International Publishing",
pages="296--308",
isbn="978-3-030-71151-1"
}

@ARTICLE{choi2009cbba,
  author={Choi, Han-Lim and Brunet, Luc and How, Jonathan P.},
  journal={IEEE Transactions on Robotics}, 
  title={Consensus-Based Decentralized Auctions for Robust Task Allocation}, 
  year={2009},
  volume={25},
  number={4},
  pages={912-926}}

@article{chovet2025performance,
  title={Performance comparison of ros2 middlewares for multi-robot mesh networks in planetary exploration},
  author={Chovet, Lo{\"\i}ck Pierre and Garcia, Gabriel Manuel and Bera, Abhishek and others},
  journal={Journal of Intelligent \& Robotic Systems},
  volume={111},
  number={1},
  pages={18},
  year={2025},
  publisher={Springer}
}

@article{chung2024learning,
  title={Learning team-based navigation: a review of deep reinforcement learning techniques for multi-agent pathfinding},
  author={Chung, Jaehoon and Fayyad, Jamil and Younes, Younes Al and Najjaran, Homayoun},
  journal={Artif. Intell. Rev.},
  volume={57},
  number={2},
  pages={41},
  year={2024},
  publisher={Springer}
}

@INPROCEEDINGS{coltin2010multi,
  author={Coltin, Brian and Liemhetcharat, Somchaya and Meriçli, Çetin and Tay, Junyun and Veloso, Manuela},
  booktitle={Proc. IEEE-RAS International Conference on Humanoid Robots}, 
  title={Multi-humanoid world modeling in Standard Platform robot soccer}, 
  year={2010},
  volume={},
  number={},
  pages={424-429}}

@inproceedings{das2019tarmac,
  title={Tarmac: Targeted multi-agent communication},
  author={Das, Abhishek and Gervet, Th{\'e}ophile and Romoff, Joshua and Batra, Dhruv and Parikh, Devi and Rabbat, Mike and Pineau, Joelle},
  booktitle={International Conference on machine learning},
  pages={1538--1546},
  year={2019},
  organization={PMLR}
}

@article{rutili2024motionplanning,
  title={Humanoid robot motion planning approaches: a survey},
  author={de Lima, Carolina Rutili and Khan, Said G and Tufail, Muhammad and Shah, Syed H and Maximo, Marcos ROA},
  journal={Journal of Intelligent \& Robotic Systems},
  volume={110},
  number={2},
  pages={86},
  year={2024},
  publisher={Springer}
}

@article{ding2025worldmodelsurvey,
author = {Ding, Jingtao and Zhang, Yunke and Shang, Yu and Zhang, Yuheng and Zong, Zefang and Feng, Jie and others},
title = {Understanding World or Predicting Future? A Comprehensive Survey of World Models},
year = {2025},
issue_date = {February 2026},
publisher = {Association for Computing Machinery},
address = {New York, NY, USA},
volume = {58},
number = {3},
issn = {0360-0300},
journal = {ACM Comput. Surv.},
articleno = {57},
numpages = {38}
}

@inproceedings{
doshi2024crossformer,
title={Scaling Cross-Embodied Learning: One Policy for Manipulation, Navigation, Locomotion and Aviation},
author={Ria Doshi and Homer Rich Walke and Oier Mees and Sudeep Dasari and Sergey Levine},
booktitle={8th Annual Conference on Robot Learning},
year={2024}
}

@inproceedings{dragan2015motion,
author = {Dragan, Anca D. and Bauman, Shira and Forlizzi, Jodi and Srinivasa, Siddhartha S.},
title = {Effects of Robot Motion on Human-Robot Collaboration},
year = {2015},
isbn = {9781450328838},
publisher = {ACM},
booktitle = {Proc. of the ACM/IEEE International Conference on Human-Robot Interaction},
pages = {51–58},
numpages = {8}
}

@INPROCEEDINGS{dragan2013legibility,
  author={Dragan, Anca D. and Lee, Kenton C.T. and Srinivasa, Siddhartha S.},
  booktitle={Proc. ACM/IEEE International Conference on Human-Robot Interaction (HRI)}, 
  title={Legibility and predictability of robot motion}, 
  year={2013},
  volume={},
  number={},
  pages={301-308}}

@INPROCEEDINGS{du2025cola,
  author={Du, Yushi and Li, Yixuan and Jia, Baoxiong and Lin, Yutang and Zhou, Pei and Liang, Wei and Yang, Yanchao and Huang, Siyuan},
  booktitle={Proc. IEEE ICRA}, 
  title={COLA: Learning Human-Humanoid Coordination for Collaborative Object Carrying}, 
  year={2026},
  volume={},
  number={},
  pages={18578-18585}}

@article{ehtesham2025survey,
  title={A survey of agent interoperability protocols: Model context protocol (mcp), agent communication protocol (acp), agent-to-agent protocol (a2a), and agent network protocol (anp)},
  author={Ehtesham, Abul and Singh, Aditi and Gupta, Gaurav Kumar and Kumar, Saket},
  journal={Preprint arXiv:2505.02279},
  year={2025}
}

@INPROCEEDINGS{otaconsensus2024,
  author={Epp, Michael and Molinari, Fabio and Raisch, Jörg},
  booktitle={Proc. IEEE 63rd Conference on Decision and Control (CDC)}, 
  title={Exploiting Over-The-Air Consensus for Collision Avoidance and Formation Control in Multi-Agent Systems}, 
  year={2024},
  volume={},
  number={},
  pages={5417-5423}}

@article{embodiedmasurvey2025,
  title={Multi-agent embodied AI: Advances and future directions},
  author={Feng, Zhaohan and Xue, Ruiqi and Yuan, Lei and Yu, Yang and Ding, Ning and Liu, Meiqin and Gao, Bingzhao and Sun, Jian and others},
  journal={Science China Information Sciences},
  volume={69},
  number={5},
  pages={151202},
  year={2026},
  publisher={Springer}
}

@inproceedings{
fu2024humanplus,
title={HumanPlus: Humanoid Shadowing and Imitation from Humans},
author={Zipeng Fu and Qingqing Zhao and Qi Wu and Gordon Wetzstein and Chelsea Finn},
booktitle={Proc. 8th Annual Conference on Robot Learning},
year={2024},
}

@article{gambo2025systematic,
  title={A systematic literature review of DDS middleware in robotic systems},
  author={Gambo, Muhammad Liman and Danasabe, Abubakar and Almadani, Basem and Aliyu, Farouq and Aliyu, Abdulrahman and Al-Nahari, Esam},
  journal={Robotics},
  volume={14},
  number={5},
  pages={63},
  year={2025},
  publisher={MDPI}
}

@inproceedings{
dreamdojo2026,
title={DreamDojo: A Generalist Robot World Model from Large-Scale Human Videos},
author={Shenyuan Gao and William Liang and Kaiyuan Zheng and Ayaan Naveed Malik and others},
booktitle={Proc. International Conference on Machine Learning},
year={2026}
}

@inproceedings{
garg2025foundation,
title={Foundation Models to the Rescue: Deadlock Resolution in Connected Multi-Robot Systems},
author={Kunal Garg and Songyuan Zhang and others},
booktitle={ICRA:  Workshop on Foundation Models and Neuro-Symbolic AI for Robotics},
year={2025}
}

@INPROCEEDINGS{gasbarrone2024decentralized,
  author={Gasbarrone, Greta and Scianca, Nicola and Lanari, Leonardo and Oriolo, Giuseppe},
  booktitle={Proc. IEEE-RAS International Conference on Humanoid Robots}, 
  title={A decentralized cooperative transportation scheme for humanoid robots}, 
  year={2024},
  volume={},
  number={},
  pages={381-387}}

@article{gerkey2004formal,
  title={A formal analysis and taxonomy of task allocation in multi-robot systems},
  author={Gerkey, Brian P and Matari{\'c}, Maja J},
  journal={The International journal of robotics research},
  volume={23},
  number={9},
  pages={939--954},
  year={2004},
  publisher={SAGE Publications}
}

@article{gielis2022critical,
  title={A critical review of communications in multi-robot systems},
  author={Gielis, Jennifer and Shankar, Ajay and Prorok, Amanda},
  journal={Current robotics reports},
  volume={3},
  number={4},
  pages={213--225},
  year={2022},
  publisher={Springer}
}

@ARTICLE{gu2025mrcographs,
  author={Gu, Qiuyi and Ye, Zhaocheng and Yu, Jincheng and Tang, Jiahao and others},
  journal={IEEE Robotics and Automation Letters}, 
  title={MR-COGraphs: Communication-Efficient Multi-Robot Open-Vocabulary Mapping System via 3D Scene Graphs}, 
  year={2025},
  volume={10},
  number={6},
  pages={5713-5720}}

@article{humanoidgym2024,
  title={Humanoid-gym: Reinforcement learning for humanoid robot with zero-shot sim2real transfer},
  author={Gu, Xinyang and Wang, Yen-Jen and Chen, Jianyu},
  journal={Preprint arXiv:2404.05695},
  year={2024}
}

@inproceedings{llmmasurvey2024,
author = {Guo, Taicheng and Chen, Xiuying and Wang, Yaqi and Chang, Ruidi and Pei, Shichao and others},
title = {Large language model based multi-agents: a survey of progress and challenges},
year = {2024},
isbn = {978-1-956792-04-1},
booktitle = {Proc. of Int. Jt. Conf. Artif. Intell.},
articleno = {890},
numpages = {10}
}

@article{hafner2025dreamerv3,
  title={Mastering diverse control tasks through world models},
  author={Hafner, Danijar and Pasukonis, Jurgis and Ba, Jimmy and Lillicrap, Timothy},
  journal={Nature},
  volume={640},
  number={8059},
  pages={647--653},
  year={2025},
  publisher={Nature Publishing Group UK London}
}

@article{hawley2019cooptransport,
  title={Control framework for cooperative object transportation by two humanoid robots},
  author={Hawley, Louis and Suleiman, Wael},
  journal={Robotics and Autonomous Systems},
  volume={115},
  pages={1--16},
  year={2019},
  publisher={Elsevier}
}

@inproceedings{
he2024omnih2o,
title={OmniH2O: Universal and Dexterous Human-to-Humanoid Whole-Body Teleoperation and Learning},
author={Tairan He and Zhengyi Luo and Xialin He and Wenli Xiao and Chong Zhang and others},
booktitle={8th Annual Conference on Robot Learning},
year={2024}
}

@INPROCEEDINGS{he2024h2o,
  author={He, Tairan and Luo, Zhengyi and Xiao, Wenli and Zhang, Chong and Kitani, Kris and Liu, Changliu and Shi, Guanya},
  booktitle={Proc. IEEE/RSJ Int. Conf. Intell. Robots Syst.}, 
  title={Learning Human-to-Humanoid Real-Time Whole-Body Teleoperation}, 
  year={2024},
  volume={},
  number={},
  pages={8944-8951}}

@INPROCEEDINGS{bayesianintent_hrc2024,
  author={Hernandez-Cruz, Vanessa and Zhang, Xiaotong and Youcef-Toumi, Kamal},
  booktitle={Proc. IEEE International Conference on Automation Science and Engineering (CASE)}, 
  title={Bayesian Intention for Enhanced Human Robot Collaboration}, 
  year={2025},
  volume={},
  number={},
  pages={1024-1030}}

@ARTICLE{4118465,
  author={Hespanha, Joo P. and Naghshtabrizi, Payam and Xu, Yonggang},
  journal={Proceedings of the IEEE}, 
  title={A Survey of Recent Results in Networked Control Systems}, 
  year={2007},
  volume={95},
  number={1},
  pages={138-162}}

@article{hoffman2024intent,
  title={Inferring human intent and predicting human action in human--robot collaboration},
  author={Hoffman, Guy and Bhattacharjee, Tapomayukh and Nikolaidis, Stefanos},
  journal={Annu. Rev. Control Robot. Auton. Syst.},
  volume={7},
  number={1},
  pages={73--95},
  year={2024},
  publisher={Annual Reviews}
}

@ARTICLE{hong2019cloudswarm,
  author={Hong, Zicong and Huang, Huawei and Guo, Song and Chen, Wuhui and Zheng, Zibin},
  journal={IEEE Transactions on Vehicular Technology}, 
  title={QoS-Aware Cooperative Computation Offloading for Robot Swarms in Cloud Robotics}, 
  year={2019},
  volume={68},
  number={4},
  pages={4027-4041}}

@inproceedings{ifl2023,
  title={Fleet-dagger: Interactive robot fleet learning with scalable human supervision},
  author={Hoque, Ryan and Chen, Lawrence Yunliang and Sharma, Satvik and Dharmarajan, Karthik and others},
  booktitle={Proc. Conference on Robot Learning},
  pages={368--380},
  year={2023},
  organization={PMLR}
}

@ARTICLE{huang2022recslam,
  author={Huang, Peng and Zeng, Liekang and Chen, Xu and Luo, Ke and Zhou, Zhi and Yu, Shuai},
  journal={IEEE Internet Things J.}, 
  title={Edge Robotics: Edge-Computing-Accelerated Multirobot Simultaneous Localization and Mapping}, 
  year={2022},
  volume={9},
  number={15},
  pages={14087-14102}}

@article{propagunsafe2026,
  title={Propagating Unsafe Actions in LLM Controlled Multi-Robot Collaboration via Single Robot Compromise},
  author={Huang, Zhen and Liu, Zhihuang and Luo, Mengxuan and Wu, Weishang and Cai, Zhiping},
  journal={Preprint arXiv:2605.15641},
  year={2026}
}

@INPROCEEDINGS{huang2025compositional,
  author={Huang, Zhehui and Shi, Guangyao and Wu, Yuwei and others},
  booktitle={Proc. IEEE International Symposium on Multi-Robot and Multi-Agent Systems (MRS)}, 
  title={Compositional Coordination for Multi-Robot Teams with Large Language Models}, 
  year={2025},
  volume={},
  number={},
  pages={}}

@article{ijaz2026jcco,
  title={JCCO: Jointly Optimizing the Computational and Communication Costs for Resource Allocation in Energy-Efficient Swarm Robotics},
  author={Ijaz, Amir and Haghbayan, Hashem and others},
  journal={Journal of Sensor and Actuator Networks},
  volume={15},
  number={4},
  pages={56},
  year={2026},
  publisher={MDPI}
}

@INPROCEEDINGS{exbody2_2024,
  author={Ji, Mazeyu and Peng, Xuanbin and Liu, Fangchen and Li, Jialong and Yang, Ge and Cheng, Xuxin and Wang, Xiaolong},
  booktitle={Proc. IEEE ICRA}, 
  title={ExBody2: Advanced Expressive Humanoid Whole-Body Control}, 
  year={2026},
  volume={},
  number={},
  pages={613-620}}

@inproceedings{jiang2025survey,
  title={A survey on vision-language-action models for autonomous driving},
  author={Jiang, Sicong and Huang, Zilin and Qian, Kangan and Luo, Ziang and Zhu, Tianze  and others},
  booktitle={Proc. of the IEEE/CVF Int. Conf. Comput. Vis. Workshop},
  pages={4583--4595},
  year={2025}
}

@ARTICLE{fedpoison2024,
  author={Junior, Anthony Jacklingo Kwame Quansah and Tutu Tchao, Eric and Al-Khalidi, Mohammed and others},
  journal={IEEE Access}, 
  title={A Systematic Review on the Practicality of Poisoning Defenses in Federated IoT Systems}, 
  year={2025},
  volume={13},
  number={},
  pages={211109-211137}}

@INPROCEEDINGS{kannan2024smartllm,
  author={Kannan, Shyam Sundar and Venkatesh, Vishnunandan L. N. and Min, Byung-Cheol},
  booktitle={Proc. IEEE/RSJ Int. Conf. Intell. Robots Syst. (IROS)}, 
  title={SMART-LLM: Smart Multi-Agent Robot Task Planning using Large Language Models}, 
  year={2024},
  volume={},
  number={},
  pages={12140-12147}}

@article{mrscollabtaxonomy2026,
  title={Collaboration in multi-robot systems: Taxonomy and survey of frameworks for collaboration},
  author={Karam, Riwa and Nguyen, Alexander A and Lin, Ruoyu and Martin, David R and Morales, Diana and others},
  journal={Preprint arXiv:2603.23898},
  year={2026}
}

@INPROCEEDINGS{kaul2012real,
  author={Kaul, Sanjit and others},
  booktitle={Proc. IEEE INFOCOM}, 
  title={Real-time status: How often should one update?}, 
  year={2012},
  volume={},
  number={},
  pages={2731-2735}}

@ARTICLE{11164279,
  author={Kawaharazuka, Kento and Oh, Jihoon and Yamada, Jun and Posner, Ingmar and Zhu, Yuke},
  journal={IEEE Access}, 
  title={Vision-Language-Action Models for Robotics: A Review Towards Real-World Applications}, 
  year={2025},
  volume={13},
  number={},
  pages={162467-162504}}

@ARTICLE{cloudrobotics2015,
  author={Kehoe, Ben and Patil, Sachin and Abbeel, Pieter and Goldberg, Ken},
  journal={IEEE Transactions on Automation Science and Engineering}, 
  title={A Survey of Research on Cloud Robotics and Automation}, 
  year={2015},
  volume={12},
  number={2},
  pages={398-409}}

@inproceedings{khazatsky2024droid,
  title     = {{DROID}: A Large-Scale In-The-Wild Robot Manipulation Dataset},
  author    = {Alexander Khazatsky and Karl Pertsch and Suraj Nair
               and others},
  booktitle = {Proceedings of Robotics: Science and Systems (RSS)},
  year      = {2024},
  address   = {Delft, Netherlands}
}

@inproceedings{
kim2019schednet,
title={Learning to Schedule Communication in Multi-agent Reinforcement Learning},
author={Daewoo Kim and others},
booktitle={Proc. ICLR},
year={2019}
}

@InProceedings{kim2024openvla,
  title = 	 {OpenVLA: An Open-Source Vision-Language-Action Model},
  author =       {Kim, Moo Jin and Pertsch, Karl and Karamcheti, Siddharth and Xiao, Ted and Balakrishna, Ashwin and others},
  booktitle = 	 {Proc. of The 8th Conference on Robot Learning},
  pages = 	 {2679--2713},
  year = 	 {2025},
  volume = 	 {270},
  month = 	 {06--09 Nov},
  publisher =    {PMLR}
}

@INPROCEEDINGS{beliefstate_hrteam2025,
  author={Kolb, Jack and Feigh, Karen M.},
  booktitle={Proc. IEEE/RSJ International Conference on Intelligent Robots and Systems (IROS)}, 
  title={Inferring Belief States in Partially-Observable Human-Robot Teams}, 
  year={2024},
  volume={},
  number={},
  pages={7115-7122}}

@article{korsah2013comprehensive,
  title={A comprehensive taxonomy for multi-robot task allocation},
  author={Korsah, G Ayorkor and Stentz, Anthony and Dias, M Bernardine},
  journal={The International journal of robotics research},
  volume={32},
  number={12},
  pages={1495--1512},
  year={2013},
  publisher={SAGE Publications Sage UK: London, England}
}

@INPROCEEDINGS{kronauer2021latency,
  author={Kronauer, Tobias and Pohlmann, Joshwa and Matthé, Maximilian and others},
  booktitle={Proc. IEEE International Conference on Multisensor Fusion and Integration for Intelligent Systems (MFI)}, 
  title={Latency Analysis of ROS2 Multi-Node Systems}, 
  year={2021},
  volume={},
  number={},
  pages={1-7}}

@INPROCEEDINGS{8624927,
  author={Kumagai, Iori and others},
  booktitle={Proc. IEEE-RAS International Conference on Humanoid Robots}, 
  title={Efficient Locomotion Planning for a Humanoid Robot with Whole-Body Collision Avoidance Guided by Footsteps and Centroidal Sway Motion}, 
  year={2018},
  volume={},
  number={},
  pages={}}

@article{lagomarsino2026role,
  title={Intuitive programming, adaptive task planning, and dynamic role allocation in human--robot collaboration},
  author={Lagomarsino, Marta and Merlo, Elena and others},
  journal={Annual Review of Control, Robotics, and Autonomous Systems},
  volume={9},
  year={2025},
  publisher={Annual Reviews}
}

@article{lawrence2025norms,
author = {Lawrence, Steven and Jouaiti, Melanie and Hoey, Jesse and others},
title = {The Role of Social Norms in Human–Robot Interaction: A Systematic Review},
year = {2025},
volume = {14},
number = {3},
journal = {ACM Trans. J. Hum.-Robot Interact.},
articleno = {56},
numpages = {}
}

@article{phuma2025,
  title={PHUMA: Physically Reliable Humanoid Locomotion Dataset},
  author={Lee, Kyungmin and others},
  journal={ArXiv:2510.26236},
  year={2025}
}

@article{li2026llmmrs,
  title={Large language models for multi-robot systems: A survey},
  author={Li, Peihan and others},
  journal={Autonomous Robots},
  volume={50},
  number={3},
  pages={30},
  year={2026},
  publisher={Springer}
}

@article{embodiedaisafety2026,
  title={Safety in Embodied AI: A Survey of Risks, Attacks, and Defenses},
  author={Li, Xiao and others},
  journal={Preprint arXiv:2605.02900},
  year={2026}
}

@article{hetformation_drl2026,
  title={High-Precision Formation Control for Heterogeneous Multi-Robot Systems via Hierarchical Hybrid Physics-Informed Deep Reinforcement Learning},
  author={Li, Yanzhou and Chen, Guangli and others},
  journal={Preprint arXiv:2607.03512},
  year={2026}
}

@InProceedings{li2022star,
  title = 	 {Multi-Robot Scene Completion: Towards Task-Agnostic Collaborative Perception},
  author =       {Li, Yiming and Zhang, Juexiao and Ma, Dekun and Wang, Yue and Feng, Chen},
  booktitle = 	 {Proc. of The Conference on Robot Learning},
  pages = 	 {2062--2072},
  year = 	 {2023},
  volume = 	 {205},
  month = 	 {14--18 Dec},
  publisher =    {PMLR}
}

@INPROCEEDINGS{berkeleyhumanoid2024,
  author={Liao, Qiayuan and Zhang, Bike and Huang, Xuanyu and Huang, Xiaoyu and Li, Zhongyu and Sreenath, Koushil},
  booktitle={Proc. IEEE Int. Conf. Robot. Autom. (ICRA)}, 
  title={Berkeley Humanoid: A Research Platform for Learning-Based Control}, 
  year={2025},
  volume={},
  number={},
  pages={2897-2904}}

@INPROCEEDINGS{hzero2025,
  author={Lin, Yunfeng and Liu, Minghuan and Xue, Yufei and Zhou, Ming and Yu, Yong and others},
  booktitle={Proc. IEEE ICRA}, 
  title={H-Zero: Cross-Humanoid Locomotion Pretraining Enables Few-shot Novel Embodiment Transfer}, 
  year={2026},
  volume={},
  number={},
  pages={9306-9313}}

@INPROCEEDINGS{liu2025coherent,
  author={Liu, Kehui and Tang, Zixin and Wang, Dong and Wang, Zhigang and Li, Xuelong and Zhao, Bin},
  booktitle={Proc. IEEE Int. Conf. Robot. Autom.}, 
  title={COHERENT: Collaboration of Heterogeneous Multi-Robot System with Large Language Models}, 
  year={2025},
  volume={},
  number={},
  pages={10208-10214}}

@inproceedings{
liu2023reflect,
title={{REFLECT}: Summarizing Robot Experiences for Failure Explanation and Correction},
author={Zeyi Liu and Arpit Bahety and Shuran Song},
booktitle={Proc. CoRL: 2nd Workshop on Language and Robot Learning: Language as Grounding},
year={2023}
}

@article{liu2025ittakestwo,
  title={It takes two: Learning interactive whole-body control between humanoid robots},
  author={Liu, Zuhong and Ge, Junhao and Xiong, Minhao and Gu, Jiahao and Tang, Bowei and Jing, Wei and Chen, Siheng},
  journal={Preprint arXiv:2510.10206},
  year={2025}
}

@article{gentlehumanoid2025,
  title={Gentlehumanoid: Learning upper-body compliance for contact-rich human and object interaction},
  author={Lu, Qingzhou and Feng, Yao and Shi, Baiyu and Piseno, Michael and Bao, Zhenan and Liu, C Karen},
  journal={Preprint arXiv:2511.04679},
  year={2025}
}

@ARTICLE{11495231,
  author={Ma, Yueen and Song, Zixing and Zhuang, Yuzheng and Hao, Jianye and King, Irwin},
  journal={IEEE Transactions on Neural Networks and Learning Systems}, 
  title={A Survey on Vision--Language--Action Models for Embodied AI}, 
  year={2026},
  volume={37},
  number={7},
  pages={3031-3051}}

@INPROCEEDINGS{mahadevan2025gamechat,
  author={Mahadevan, Vagul and Zhang, Shangtong and Chandra, Rohan},
  booktitle={Proc. IEEE MRS}, 
  title={GameChat: Multi-LLM Dialogue for Safe, Agile, and Socially Optimal Multi-Agent Navigation in Constrained Environments}, 
  year={2025},
  volume={},
  number={},
  pages={1-7}}

@InProceedings{mahmood2019amass,
author = {Mahmood, Naureen and Ghorbani, Nima and Troje, Nikolaus F. and Pons-Moll, Gerard and Black, Michael J.},
title = {AMASS: Archive of Motion Capture As Surface Shapes},
booktitle = {Proc. of the IEEE/CVF International Conference on Computer Vision (ICCV)},
month = {October},
year = {2019}
}

@inproceedings{
makoviychuk2021isaacgym,
title={Isaac Gym: High Performance {GPU} Based Physics Simulation For Robot Learning},
author={Viktor Makoviychuk and Lukasz Wawrzyniak and others},
booktitle={Proc. Conference on Neural Information Processing Systems Datasets and Benchmarks Track},
year={2021}
}

@INPROCEEDINGS{mandi2024roco,
  author={Mandi, Zhao and Jain, Shreeya and Song, Shuran},
  booktitle={Proc. IEEE International Conference on Robotics and Automation (ICRA)}, 
  title={RoCo: Dialectic Multi-Robot Collaboration with Large Language Models}, 
  year={2024},
  volume={},
  number={},
  pages={286-299}}

@inproceedings{maruyama2016exploring,
author = {Maruyama, Yuya and Kato, Shinpei and Azumi, Takuya},
title = {Exploring the performance of ROS2},
year = {2016},
isbn = {9781450344852},
publisher = {Association for Computing Machinery},
articleno = {5},
numpages = {10},
location = {Pittsburgh, Pennsylvania},
booktitle = {Proc. EMSOFT}
}

@INPROCEEDINGS{mcgill2011stretcher,
  author={McGill, Stephen G. and Lee, Daniel D.},
  booktitle={2011 11th IEEE-RAS International Conference on Humanoid Robots}, 
  title={Cooperative humanoid stretcher manipulation and locomotion}, 
  year={2011},
  volume={},
  number={},
  pages={429-433}}

@inproceedings{ghosh2024octo,
    title={Octo: An Open-Source Generalist Robot Policy},
    author = {{Octo Model Team} and Dibya Ghosh and Homer Walke and Karl Pertsch and Kevin Black and Oier Mees and others},
    booktitle = {Proceedings of Robotics: Science and Systems},
    address  = {Delft, Netherlands},
    year = {2024},
}

@article{mehrotra2024appropriatetrust,
author = {Mehrotra, Siddharth and Degachi, Chadha and others},
title = {A Systematic Review on Fostering Appropriate Trust in Human-AI Interaction: Trends, Opportunities and Challenges},
year = {2024},
publisher = {ACM},
volume = {1},
number = {4},
journal = {ACM J. Responsib. Comput.},
articleno = {26}
}

@article{safefall2025,
  title={Safefall: Learning protective control for humanoid robots},
  author={Meng, Ziyu and Liu, Tengyu and Ma, Le and Wu, Yingying and Song, Ran and Zhang, Wei and Huang, Siyuan},
  journal={Preprint arXiv:2511.18509},
  year={2025}
}

@article{mittal2025isaaclab,
  title={Isaac lab: A gpu-accelerated simulation framework for multi-modal robot learning},
  author={Mittal, Mayank and Roth, Pascal and Tigue, James and Richard, Antoine and Zhang, Octi and Du, Peter and Serrano-Munoz, Antonio and others},
  journal={Preprint arXiv:2511.04831},
  year={2025}
}

@article{montano2016coordination,
title = {Coordination of several robots based on temporal synchronization},
journal = {Robotics and Computer-Integrated Manufacturing},
volume = {42},
pages = {73-85},
year = {2016},
issn = {0736-5845},
author = {Andrés Montaño and Raúl Suárez},
}

@INPROCEEDINGS{mox2024opportunistic,
  author={Mox, Daniel and Garg, Kashish and Ribeiro, Alejandro and Kumar, Vijay},
  booktitle={Proc. IEEE International Conference on Robotics and Automation (ICRA)}, 
  title={Opportunistic Communication in Robot Teams}, 
  year={2024},
  volume={},
  number={},
  pages={12090-12096}}

@ARTICLE{mu2026social,
  author={Mu, Xuan and Liu, Xiaorui and Fan, Qingyang and Liu, Lin and Yu, Jinpeng},
  journal={IEEE Trans. Autom. Sci. Eng.}, 
  title={A Social Navigation Framework Based on Laplace Group Dynamics for Humanoid Robot in the Dynamic Scenario}, 
  year={2026},
  volume={23},
  number={},
  pages={6698-6710}}

@INPROCEEDINGS{openbotfleet2024,
  author={Müller, Matthias and Brahmbhatt, Samarth and Deka, Ankur and Leboutet, Quentin and Hafner, David and Koltun, Vladlen},
  booktitle={Proc. IEEE Int. Conf. Robot. Autom.}, 
  title={OpenBot-Fleet: A System for Collective Learning with Real Robots}, 
  year={2024},
  volume={},
  number={},
  pages={4758-4765}}

@article{nadiri2025swarm,
  title={Swarm intelligence for collaborative play in humanoid soccer teams},
  author={Nadiri, Farzad and Rad, Ahmad B},
  journal={Sensors},
  volume={25},
  number={11},
  pages={3496},
  year={2025},
  publisher={MDPI}
}

@article{nahum2026proxemics,
  title={Advancing a taxonomy of proxemics for socially aware robot navigation},
  author={Nahum, Ehud and Edan, Yael and Oron-Gilad, Tal},
  journal={Frontiers in Robotics and AI},
  volume={13},
  pages={1800762},
  year={2026},
  publisher={Frontiers Media SA}
}

@article{neuber2026gaze,
  title={Eyes ahead: a scoping review of technologies enabling humanoid robots to follow human gaze},
  author={Neuber, Leana and Culemann, Wolf and Ingendoh, Ruth Maria and Heine, Angela},
  journal={Frontiers in Robotics and AI},
  volume={12},
  pages={1723527},
  year={2025},
  publisher={Frontiers Media SA}
}

@article{simtom_hrteams2025,
  title={Simulation theory of mind for heterogeneous human-robot teams},
  author={Nicolescu, Monica and Blankenburg, Janelle and Anima, Bashira Akter and Zagainova, Mariya and Hoseini, Pourya and others},
  journal={Frontiers in Robotics and AI},
  volume={12},
  pages={1533054},
  year={2025},
  publisher={Frontiers Media SA}
}

@inproceedings{nikolaidis2017mutual,
author = {Nikolaidis, Stefanos and Zhu, Yu Xiang and Hsu, David and Srinivasa, Siddhartha},
title = {Human-Robot Mutual Adaptation in Shared Autonomy},
year = {2017},
isbn = {9781450343367},
publisher = {ACM},
booktitle = {Proc. of ACM/IEEE International Conference on Human-Robot Interaction},
pages = {294–302},
numpages = {9},
}

@ARTICLE{1333204,
  author={Olfati-Saber, R. and Murray, R.M.},
  journal={IEEE Transactions on Automatic Control}, 
  title={Consensus problems in networks of agents with switching topology and time-delays}, 
  year={2004},
  volume={49},
  number={9},
  pages={1520-1533}}

@INPROCEEDINGS{oxe2023,
  author={O’Neill, Abby and Rehman, Abdul and Maddukuri, Abhiram and others},
  booktitle={Proc. IEEE Int. Conf. Robot. Autom.}, 
  title={Open X-Embodiment: Robotic Learning Datasets and RT-X Models: Open X-Embodiment Collaboration$^{0}$}, 
  year={2024},
  volume={},
  number={},
  pages={6892-6903}}

@ARTICLE{park2020realtime,
  author={Park, Jaeho and Delgado, Raimarius and Choi, Byoung Wook},
  journal={IEEE Access}, 
  title={Real-Time Characteristics of ROS 2.0 in Multiagent Robot Systems: An Empirical Study}, 
  year={2020},
  volume={8},
  number={},
  pages={154637-154651}}

@INPROCEEDINGS{peng2025milp,
  author={Peng, Mingming and Chen, Zhendong and Yang, Jie and others},
  booktitle={Proc. IEEE Int. Conf. Intell. Robots Syst.}, 
  title={Automatic MILP Model Construction for Multi-Robot Task Allocation and Scheduling Based on Large Language Models}, 
  year={2025},
  volume={},
  number={},
  pages={20291-20296}}

@INPROCEEDINGS{puck2021performance,
  author={Puck, L. and Keller, P. and Schnell, T. and Plasberg, C. and Tanev, A. and Heppner, G. and others},
  booktitle={Proc. IEEE Int. Conf. Autom. Sci. Eng}, 
  title={Performance Evaluation of Real-Time ROS2 Robotic Control in a Time-Synchronized Distributed Network}, 
  year={2021},
  volume={},
  number={},
  pages={1670-1676}}

@article{rahman2019offloading,
  title={Energy-efficient optimal task offloading in cloud networked multi-robot systems},
  author={Rahman, Akhlaqur and Jin, Jiong and Rahman, Ashfaqur and Cricenti, Antonio and Afrin, Mahbuba and Dong, Yu-ning},
  journal={Computer Networks},
  volume={160},
  pages={11--32},
  year={2019},
  publisher={Elsevier}
}

@article{roesler2024trustdynamics,
title = {The dynamics of human–robot trust attitude and behavior — Exploring the effects of anthropomorphism and type of failure},
journal = {Computers in Human Behavior},
volume = {150},
pages = {108008},
year = {2024},
issn = {0747-5632},
author = {Eileen Roesler and Meret Vollmann and others}
}

@INPROCEEDINGS{10375191,
  author={Rossini, Luca and Hoffman, Enrico Mingo and others},
  booktitle={Proc. IEEE-RAS International Conference on Humanoid Robots (Humanoids)}, 
  title={A Real-Time Approach for Humanoid Robot Walking Including Dynamic Obstacles Avoidance}, 
  year={2023},
  volume={},
  number={},
  pages={1-8}}

@article{rubagotti2022perceived,
title = {Perceived safety in physical human–robot interaction—A survey},
journal = {Robotics and Autonomous Systems},
volume = {151},
pages = {104047},
year = {2022},
issn = {0921-8890},
author = {Matteo Rubagotti and Inara Tusseyeva and Sara Baltabayeva and Danna Summers and Anara Sandygulova}
}

@article{saettone2026culture,
  title={Diversity and Culture in Social Robotics: A Scoping Review},
  author={Saettone, Lorenza and Sgorbissa, Antonio and Recchiuto, Carmine Tommaso},
  journal={International Journal of Social Robotics},
  volume={18},
  number={3},
  pages={51},
  year={2026},
  publisher={Springer}
}

@inproceedings{sauppe2014deictics,
author = {Saupp{\'e}, Allison and Mutlu, Bilge},
title = {Robot deictics: how gesture and context shape referential communication},
year = {2014},
isbn = {9781450326582},
booktitle = {Proc. of the 2014 ACM/IEEE International Conference on Human-Robot Interaction},
pages = {342–349},
numpages = {8},
location = {Bielefeld, Germany},
}

@article{scheutz2017shared,
  title   = {A Framework for Developing and Using Shared Mental Models in Human-Agent Teams},
  author  = {Scheutz, Matthias and DeLoach, Scott A. and Adams, Julie A.},
  journal = {Journal of Cognitive Engineering and Decision Making},
  volume  = {11},
  number  = {3},
  pages   = {203--224},
  year    = {2017}
}

@article{sebanz2006joint,
  title   = {Joint Action: Bodies and Minds Moving Together},
  author  = {Sebanz, Natalie and Bekkering, Harold and Knoblich, G{\"u}nther},
  journal = {Trends in Cognitive Sciences},
  volume  = {10},
  number  = {2},
  pages   = {70--76},
  year    = {2006}
}

@ARTICLE{seraj2024hetero,
  author={Seraj, Esmaeil and Paleja, Rohan and Pimentel, Luis and Lee, Kin Man and Wang, Zheyuan and others},
  journal={IEEE Transactions on Robotics}, 
  title={Heterogeneous Policy Networks for Composite Robot Team Communication and Coordination}, 
  year={2024},
  volume={40},
  number={},
  pages={3833-3849}}

@inproceedings{sferrazza2024humanoidbench,
  author    = {Sferrazza, Carmelo and Huang, Dun-Ming and others},
  title     = {HumanoidBench: Simulated Humanoid Benchmark for Whole-Body Locomotion and Manipulation},
  booktitle = {Proceedings of Robotics: Science and Systems (RSS)},
  year      = {2024},
  address   = {Delft, Netherlands},
  month     = {July},
}

@article{sheng2025smartbot,
author = {Sheng, Qincheng and Zhou, Zhongxiang and Li, Jinhao and Mi, Xiangyu and Xiang, Pingyu and Chen, Zhenghan and Xu, Haocheng and Jia, Shenhan and Wu, Xiyang and others},
title = {A Comprehensive Review of Humanoid Robots},
journal = {SmartBot},
volume = {1},
number = {1},
pages = {e12008},
year = {2025}
}

@ARTICLE{shorinwa2024distopt,
  author={Shorinwa, Ola and Halsted, Trevor and Yu, Javier and Schwager, Mac},
  journal={IEEE Robotics \& Automation Magazine}, 
  title={Distributed Optimization Methods for Multi-Robot Systems: Part 1—A Tutorial [Tutorial]}, 
  year={2024},
  volume={31},
  number={3},
  pages={121-138}}

@misc{getupmorph2025,
      title={Learning to Get Up Across Morphologies: Zero-Shot Recovery with a Unified Humanoid Policy}, 
      author={Jonathan Spraggett},
      year={2025},
      eprint={2512.12230},
      archivePrefix={arXiv},
      primaryClass={cs.RO}
}

@inproceedings{sukhbaatar2016commnet,
 author = {Sukhbaatar, Sainbayar and szlam, arthur and Fergus, Rob},
 booktitle = {Proc. Advances in Neural Information Processing Systems},
 pages = {},
 title = {Learning Multiagent Communication with Backpropagation},
 volume = {29},
 year = {2016}
}

@misc{multiagentcoord2025,
      title={Multi-Agent Coordination across Diverse Applications: A Survey}, 
      author={Lijun Sun and Yijun Yang and Qiqi Duan and Yuhui Shi and Chao Lyu and Yu-Cheng Chang and Chin-Teng Lin and Yang Shen},
      year={2025},
      eprint={2502.14743},
      archivePrefix={arXiv},
      primaryClass={cs.MA},
}

@inproceedings{tan1993iql,
author = {Tan, Ming},
title = {Multi-agent reinforcement learning: independent versus cooperative agents},
year = {1993},
isbn = {1558603077},
booktitle = {Proceedings of the Tenth International Conference on International Conference on Machine Learning},
pages = {330–337},
numpages = {8},
}

@article{tao2026tamp,
  title   = {Multi-Agent Task and Motion Planning Trends Analysis: A Survey},
  author  = {Tao, X. and Mart{\'i}nez-Ortega, Juan F. and Lucas-Mart{\'i}nez, N. and Mario San-Emeterio-de-la-Parte},
  journal = {Artificial Intelligence Review},
  volume  = {59},
  number  = {9},
  pages   = {198},
  year    = {2026}
}

@ARTICLE{verhagen2024meaningful,
AUTHOR={Verhagen, Ruben S.  and Neerincx, Mark A.  and Tielman, Myrthe L. }, 
TITLE={Meaningful human control and variable autonomy in human-robot teams for firefighting},
JOURNAL={Frontiers in Robotics and AI}, 
VOLUME={Volume 11 - 2024},
YEAR={2024},
ISSN={2296-9144},
}

@article{collabmind_intent2021,
  title   = {The Collaborative Mind: Intention Reading and Trust in Human-Robot Interaction},
  author  = {Vinanzi, Samuele and Cangelosi, Angelo and Goerick, Christian},
  journal = {iScience},
  volume  = {24},
  number  = {2},
  year    = {2021}
}

@INPROCEEDINGS{wagner2026races,
  author={Wagner, Niklas A. and Wietfeld, Christian},
  booktitle={Proc. IEEE Conf. Comput. Commun.}, 
  title={O-RACES: Proactive AI-driven Scheduling in Open RAN for 6G-Networked Humanoid Robots}, 
  year={2026},
  volume={},
  number={},
  pages={},
  doi={10.1109/INFOCOM59046.2026.11571197}}

@inproceedings{wang2023fleetmerge,
 author = {Wang, Lirui and Zhang, Kaiqing and Zhou, Allan and Simchowitz, Max and Tedrake, Russ},
 booktitle = {Proc. International Conference on Learning Representations},
 pages = {51737--51771},
 title = {Robot Fleet Learning via Policy Merging},
 volume = {2024},
 year = {2024}
}

@inproceedings{
lwd2026,
title={Learning While Deploying: Fleet-Scale Reinforcement Learning for Generalist Robot Policies},
author={Yi Wang and Xinchen Li and Pengwei Xie and others},
booktitle={RSS 2026 Workshop: Reinforcement Learning for Vision-Language-Action Models Workshop},
year={2026},
}

@misc{wang2025dartllm,
      title={DART-LLM: Dependency-Aware Multi-Robot Task Decomposition and Execution using Large Language Models}, 
      author={Yongdong Wang and Runze Xiao and Jun Younes Louhi Kasahara and Ryosuke Yajima and others},
      year={2025},
      eprint={2411.09022},
      archivePrefix={arXiv},
      primaryClass={cs.RO}
}

@Article{dualhumanoidtransport2024,
AUTHOR = {Wen, Shengjun and Shi, Zhaoyuan and Li, Hongjun},
TITLE = {Coordinated Transport by Dual Humanoid Robots Using Distributed Model Predictive Control},
JOURNAL = {Biomimetics},
VOLUME = {9},
YEAR = {2024},
NUMBER = {6},
ARTICLE-NUMBER = {332},
PubMedID = {38921212},
ISSN = {2313-7673},
}

@article{wong2017workload,
author = {Choon Yue Wong and Gerald Seet},
title ={Workload, awareness and automation in multiple-robot supervision},
journal = {International Journal of Advanced Robotic Systems},
volume = {14},
number = {3},
pages = {1729881417710463},
year = {2017}}

@inproceedings{wu2025generative,
author = {Wu, Di and Wei, Xian and Chen, Guang and Shen, Hao and Jin, Bo},
title = {Generative multi-agent collaboration in embodied AI: a systematic review},
year = {2025},
isbn = {978-1-956792-06-5},
booktitle = {Proc. of International Joint Conference on Artificial Intelligence},
articleno = {1190},
numpages = {},
}

@misc{wu2025survey,
      title={A Survey on Cloud-Edge-Terminal Collaborative Intelligence in AIoT Networks}, 
      author={Jiaqi Wu and Jing Liu and Yang Liu and Lixu Wang and Zehua Wang and Wei Chen and Zijian Tian and Richard Yu and Victor C. M. Leung},
      year={2025},
      eprint={2508.18803},
      archivePrefix={arXiv},
      primaryClass={cs.NI}
}

@INPROCEEDINGS{coophumanoidtransport,
  author={Meng-Hung Wu and Konno, Atsushi and Uchiyama, Masaru},
  booktitle={Proc. 2011 IEEE/SICE International Symposium on System Integration (SII)}, 
  title={Cooperative object transportation by multiple humanoid robots}, 
  year={2011},
  volume={},
  number={},
  pages={779-784}}

@article{wu2016symmetry,
author = {Meng-Hung Wu and Shuhei Ogawa and Atsushi Konno},
title = {Symmetry position/force hybrid control for cooperative object transportation using multiple humanoid robots},
journal = {Advanced Robotics},
volume = {30},
number = {2},
pages = {131--149},
year = {2016},
publisher = {Taylor \& Francis},
}

@article{pandit2026decmht,
  title={Learning Multi-Humanoid Pickup and Transport via Decentralized Object-Centric Control},
  author={Pandit, Bikram and Gadde, Mohitvishnu S and Shrestha, Aayam Kumar and Fern, Alan},
  journal={arXiv preprint arXiv:2609.17824},
  year={2026}
}

@ARTICLE{wu2024gsrc,
  author={Wu, Wenchao and Yang, Yuanqing and others},
  journal={IEEE Trans. Wirel. Commun.}, 
  title={Goal-Oriented Semantic Communications for Robotic Waypoint Transmission: The Value and Age of Information Approach}, 
  year={2024},
  volume={23},
  number={12},
  pages={18903-18915}}

@inproceedings{
yueqi2024fedredefense,
title={Fed{RED}efense: Defending against Model Poisoning Attacks for Federated Learning using Model Update Reconstruction Error},
author={Yueqi XIE and Minghong Fang and Neil Zhenqiang Gong},
booktitle={Proc. ICML},
year={2024}
}

@misc{xue2026taskoriented,
      title={From Raw Data to Shared 3D Semantics: Task-Oriented Communication for Multi-Robot Collaboration}, 
      author={Ruibo Xue and Jiedan Tan and Fang Liu and Jingwen Tong and Taotao Wang and Shuoyao Wang},
      year={2026},
      eprint={2602.08624},
      archivePrefix={arXiv},
      primaryClass={cs.NI}
}

@inproceedings{
crosshumanoidwbc2026,
title={Scalable and General Whole-Body Control for Cross-Humanoid Locomotion},
author={Yufei Xue and Yunfeng Lin and Wentao Dong and Yang Tang and Jingbo Wang and Jiangmiao Pang and others},
booktitle={Proc. International Conference on Machine Learning},
year={2026}
}

@INPROCEEDINGS{yahya2017collective,
  author={Yahya, Ali and Li, Adrian and Kalakrishnan, Mrinal and Chebotar, Yevgen and Levine, Sergey},
  booktitle={Proc. IEEE/RSJ Int. Conf. Intell. Robots Syst.}, 
  title={Collective robot reinforcement learning with distributed asynchronous guided policy search}, 
  year={2017},
  volume={},
  number={},
  pages={79-86}}

@Article{yang2025collision,
AUTHOR = {Yang, Guanghong and An, Liwei and Zhao, Can},
TITLE = {Collision/Obstacle Avoidance Coordination of Multi-Robot Systems: A Survey},
JOURNAL = {Actuators},
VOLUME = {14},
YEAR = {2025},
NUMBER = {2},
ARTICLE-NUMBER = {85},
ISSN = {2076-0825},
}

@ARTICLE{yang2026r2x,
  author={Yang, Hyun Jong and Lee, Howon and  others},
  journal={IEEE Communications Surveys \& Tutorials}, 
  title={Advancing Multi-Robot Networks via MLLM-Driven Sensing, Communication, and Computation: A Comprehensive Survey}, 
  year={2026},
  volume={28},
  number={},
  pages={5833-5871}}

@INPROCEEDINGS{cbfrl2025,
  author={Yang, Lizhi and Werner, Blake and de Sa, Massimiliano and Ames, Aaron D.},
  booktitle={Proc. IEEE ICRA}, 
  title={CBF-RL: Safety Filtering Reinforcement Learning in Training with Control Barrier Functions}, 
  year={2026},
  volume={},
  number={},
  pages={}}

@ARTICLE{9380899,
  author={Yates, Roy D. and Sun, Yin and Brown, D. Richard and Kaul, Sanjit K. and Modiano, Eytan and Ulukus, Sennur},
  journal={IEEE Journal on Selected Areas in Communications}, 
  title={Age of Information: An Introduction and Survey}, 
  year={2021},
  volume={39},
  number={5},
  pages={1183-1210}}

@article{yi2016hierarchical,
author = {Seung-Joon Yi and Stephen McGill and Dennis Hong and Daniel Lee},
title ={Hierarchical Motion Control for a Team of Humanoid Soccer Robots},
journal = {International Journal of Advanced Robotic Systems},
volume = {13},
number = {1},
pages = {32},
year = {2016},
}

@article{you2023subgroup,
  title   = {Subgroup Formation in Human--Robot Teams: A Multi-Study Mixed-Method Approach with Implications for Theory and Practice},
  author  = {You, Sangseok and Robert, Lionel P.},
  journal = {J. Assoc. Inf. Sci. Technol.},
  volume  = {74},
  number  = {3},
  pages   = {323--338},
  year    = {2023}
}

@ARTICLE{yu2026ioh,
  author={Yu, Angela W. and Nayak, Amiya},
  journal={IEEE Internet of Things Journal}, 
  title={The Internet of Humanoids: A Survey of Technologies, Applications, and Challenges}, 
  year={2026},
  volume={13},
  number={6},
  pages={10498-10521}}

@ARTICLE{yuan2025survey,
  author={Yuan, Mingqi and Yu, Tao and Ge, Wenqi and Yao, Xiuyong and others},
  journal={IEEE Trans. Pattern Anal. Mach. Intell.}, 
  title={A Survey of Behavior Foundation Model: Next-Generation Whole-Body Control System of Humanoid Robots}, 
  year={2026},
  volume={48},
  number={4},
  pages={4909-4927}}

@article{zhai2021cloudroid,
  title   = {Cloudroid Swarm: A QoS-Aware Framework for Multirobot Cooperation Offloading},
  author  = {Zhai, Yuanzhao and Ding, Bo and Zhang, Pengfei and Luo, Jie},
  journal = {Wireless Communications and Mobile Computing},
  volume  = {2021},
  number  = {1},
  pages   = {6631111},
  year    = {2021}
}

@inproceedings{coela2023,
 author = {Zhang, Hongxin and Du, Weihua and Shan, Jiaming and Zhou, Qinhong and Du, Yilun and Tenenbaum, Joshua B and others},
 booktitle = {Proc. ICLR},
 pages = {19373--19401},
 title = {Building Cooperative Embodied Agents Modularly with Large Language Models},
 volume = {2024},
 year = {2024}
}

@inproceedings{combo2025,
 author = {Zhang, Hongxin and Wang, Zeyuan and Lyu, Qiushi and others},
 booktitle = {Proc. International Conference on Learning Representations},
 pages = {49996--50019},
 title = {COMBO: Compositional World Models for Embodied Multi-Agent Cooperation},
 volume = {2025},
 year = {2025}
}

@article{zhang2024middleware,
  title   = {Comparison of Middlewares in Edge-to-Edge and Edge-to-Cloud Communication for Distributed {ROS} 2 Systems},
  author  = {Zhang, Jiaqiang and Yu, Xianjia and Ha, Sier and others},
  journal = {Journal of Intelligent \& Robotic Systems},
  volume  = {110},
  number  = {4},
  pages   = {162},
  year    = {2024}
}

@misc{worldmodelsurvey2025,
      title={A Step Toward World Models: A Survey on Robotic Manipulation}, 
      author={Peng-Fei Zhang and Ying Cheng and Xiaofan Sun and Shijie Wang and Fengling Li and Lei Zhu and Heng Tao Shen},
      year={2025},
      eprint={2511.02097},
      archivePrefix={arXiv},
      primaryClass={cs.RO}
}

@article{multirobotsurvey2025,
  title   = {A Survey on Multi-Robot Collaboration Systems: Architectures, Performances, and Applications},
  author  = {Zhang, Simon and Li, Zhengxiong and Yin, Yaxuan and Xu, Shuai and Chaudhary, Vipin and Xu, Hailu},
  journal = {TechRxiv Preprints},
  year    = {2025}
}

@article{zhang2025tsn,
author = {Zhang, Tianyu and Wang, Gang and Xue, Chuanyu and Wang, Jiachen and others},
title = {Time-Sensitive Networking (TSN) for Industrial Automation: Current Advances and Future Directions},
year = {2024},
publisher = {ACM},
volume = {57},
number = {2},
issn = {0360-0300},
journal = {ACM Comput. Surv.},
articleno = {30}
}

@Article{zhao2026desync,
AUTHOR = {Zhao, Rong and Que, Haohua and Sun, Jiajun and Xie, Jiayue and Zhang, Qian and Qiao, Fei},
TITLE = {Don’t Fire Together: Desynchronised Communication Scheduling for Bandwidth-Limited Multi-Robot Exploration},
JOURNAL = {Sensors},
VOLUME = {26},
YEAR = {2026},
NUMBER = {15},
ARTICLE-NUMBER = {4778},
PubMedID = {42590554},
ISSN = {1424-8220}
}

@INPROCEEDINGS{humanoideveryday2025,
  author={Zhao, Zhenyu and Jing, Hongyi and Liu, Xiawei and Mao, Jiageng and Jha, Abha and Yang, Hanwen and Xue, Rong and others},
  booktitle={Proc. IEEE ICRA}, 
  title={Humanoid Everyday: A Comprehensive Robotic Dataset for Open-World Humanoid Manipulation}, 
  year={2026},
  volume={},
  number={},
  pages={6415-6422}}

@article{zhou2021review,
  title   = {Multi-Robot Coordination and Planning in Uncertain and Adversarial Environments},
  author  = {Zhou, Lifeng and Tokekar, Pratap},
  journal = {Current Robotics Reports},
  volume  = {2},
  number  = {2},
  pages   = {147--157},
  year    = {2021}
}

@INPROCEEDINGS{zhu2020remotewbc,
  author={Zhu, Huaijiang and Sharma, Manali and Pfeiffer, Kai and Mezzavilla, Marco and Shen, Jia and others},
  booktitle={Proc. IEEE/RSJ Int. Conf. Intell. Robots Syst. (IROS)}, 
  title={Enabling Remote Whole-Body Control with 5G Edge Computing}, 
  year={2020},
  volume={},
  number={},
  pages={3553-3560}}

@InProceedings{brohan2023rt2,
  title = 	 {RT-2: Vision-Language-Action Models Transfer Web Knowledge to Robotic Control},
  author =       {Zitkovich, Brianna and Yu, Tianhe and Xu, Sichun and Xu, Peng and Xiao, Ted and others},
  booktitle = 	 {Proc. of Conference on Robot Learning},
  pages = 	 {2165--2183},
  year = 	 {2023},
  volume = 	 {229},
  month = 	 {06--09 Nov},
  publisher =    {PMLR},
}

@ARTICLE{cap2015prioritized,
  author={Čáp, Michal and Novák, Peter and Kleiner, Alexander and Selecký, Martin},
  journal={IEEE Trans. Autom. Sci. Eng.}, 
  title={Prioritized Planning Algorithms for Trajectory Coordination of Multiple Mobile Robots}, 
  year={2015},
  volume={12},
  number={3},
  pages={835-849}}

@article{muralidharan2021communication,
   author = "Muralidharan, Arjun and Mostofi, Yasamin",
   title = "Communication-Aware Robotics: Exploiting Motion for Communication", 
   journal= "Annual Review of Control, Robotics, and Autonomous Systems",
   year = "2021",
   volume = "4",
   number = "Volume 4, 2021",
   pages = "115-139",
   publisher = "Annual Reviews",
   issn = "2573-5144",
   type = "Journal Article",
}

@misc{1x2026neo,
  author       = {{1X Technologies}},
  title        = {{NEO Home Robot}},
  year         = {2026},
  howpublished = {\url{https://www.1x.tech/neo}},
  note         = {Accessed: 2026-09-19}
}

@misc{figure2025bmw,
  author       = {{Figure AI}},
  title        = {{F.02 Contributed to the Production of 30,000 Cars at BMW}},
  year         = {2025},
  month        = nov,
  howpublished = {\url{https://www.figure.ai/news/production-at-bmw}},
  note         = {Accessed: 2026-09-19}
}

@misc{hrt2026top12,
  author       = {{Humanoid Robotics Technology}},
  title        = {Top 12 Humanoid Robots of 2026},
  year         = {2026},
  month        = jan,
  howpublished = {\url{https://humanoidroboticstechnology.com/articles/top-12-humanoid-robots-of-2026/}},
  note         = {Accessed: July 17, 2026}
}

@misc{therobotreport2026optimus,
  author       = {{The Robot Report}},
  title        = {From EVs to Robotics: Tesla Targets 10M Optimus Units with New Texas Plant},
  year         = {2026},
  month        = apr,
  howpublished = {\url{https://www.therobotreport.com/from-evs-to-robotics-tesla-targets-10m-optimus-units-with-new-texas-plant/}}
}

@misc{figure2025helix,
  author = {{Figure AI}},
  title = {Helix: A Vision-Language-Action Model for Generalist Humanoid Control},
  howpublished = {\url{https://www.figure.ai/news/helix}},
  year = {2025},
}

@ARTICLE{11192484,
  author={Nguyen, Loc X. and Raha, Avi Deb and others},
  journal={IEEE Commun. Surv. Tutor.}, 
  title={A Contemporary Survey on Semantic Communications: Theory of Mind, Generative AI, and Deep Joint Source-Channel Coding}, 
  year={2026},
  volume={28},
  number={},
  pages={2377-2417}}

\end{document}